%% file: acl_latex.tex
\documentclass[11pt]{article}

\usepackage[final]{acl}

\usepackage{times}
\usepackage{latexsym}
\usepackage{booktabs} 
\usepackage{amssymb}   % 提供 \checkmark
\usepackage{pifont}    % 提供 \ding{55}

\usepackage[T1]{fontenc}
\usepackage[utf8]{inputenc}

\usepackage{microtype}

\usepackage{inconsolata}

\usepackage{graphicx}
\usepackage{hyperref}       % hyperlinks
\usepackage{url}            % simple URL typesetting
\usepackage{booktabs}       % professional-quality tables
\usepackage{amsfonts}       % blackboard math symbols
\usepackage{nicefrac}       % compact symbols for 1/2, etc.
\usepackage{microtype}      % microtypography
\usepackage{xcolor}         % colors
\usepackage{amsmath}
\usepackage{multirow}
\usepackage{graphicx} 
\usepackage{dsfont}
\usepackage{tikz}

\usepackage{enumitem}
\usepackage{pifont}
\usepackage{caption}
\usepackage{amssymb}
\usepackage{wrapfig}
\usepackage{bm}
\usepackage{subcaption}

\usepackage{algorithm}
\usepackage{algpseudocode}

\usepackage[most]{tcolorbox}
\usepackage{fontawesome5} 
\usepackage{enumitem}
\newtcolorbox{promptbox}[1][]{
    colback=gray!10!white,
    colframe=gray!60!black,
    title=Prompt,
    coltitle=white,
    fonttitle=\bfseries,
    #1
}
\title{Beyond Solvability: Task Learnability as a Static Prior for LLM RL Post-Training}

\author{
  \textbf{Ting Zhou\textsuperscript{1}\thanks{Equal contribution.}},
  \textbf{Zhenqing Ling\textsuperscript{2}\footnotemark[1]},
  \textbf{Daoyuan Chen\textsuperscript{2}},
  \textbf{Qianli Shen\textsuperscript{2}},
  \textbf{Yilun Huang\textsuperscript{2}}, \\
  \textbf{Ying Shen\textsuperscript{1}\thanks{Corresponding author.}},
  \textbf{Yaliang Li\textsuperscript{2}} \\
  \textsuperscript{1}Sun Yat-Sen University \quad
  \textsuperscript{2}Alibaba Group \\
  \small
  zhout88@mail2.sysu.edu.cn, sheny76@mail.sysu.edu.cn \\
  \small
  lingzhenqing.lzq, daoyuanchen.cdy, shenqianli.sql, lielin.hyl, yaliang.li @alibaba-inc.com
}

\begin{document}
\maketitle
\input{section/00_abstract}
\input{section/01_intro}
\input{section/02_related_works}

\input{section/03_learnability}

\input{section/04_method}

\input{section/05_exp}
\input{section/06_conclusion}


\bibliography{custom}

\appendix
\input{section/Appendix/implementation}

% \clearpage
\input{section/Appendix/extended_exps}
% \clearpage
\input{section/Appendix/complete_res}
% \clearpage

\input{section/Appendix/overall_algorithm}

% \section{Example Appendix}
% \label{sec:appendix}

% This is an appendix.

\end{document}

%% file: section/00_abstract.tex
\begin{abstract}
Reinforcement learning (RL) has become a central post-training paradigm for eliciting reasoning capabilities in large language models, yet uniform task sampling allocates compute without regard to differences in how tasks respond to optimization. 
Existing task-valuation methods mostly rely on snapshot-based signals such as current pass rate or reward, which estimate how solvable a task is under the current policy. However, tasks with similar current solvability can still differ substantially in how positively they respond to further training.
We study this residual axis as \emph{task learnability}: a regime-conditional measure of expected positive response to continued training under a fixed RL post-training regime. By analyzing per-task reward trajectories, we find that learnability is reproducible across independently sampled training contexts and predictive of downstream utility. To make this signal practical before training begins, we propose \textsc{TrajVal}, a lightweight probe-based estimator that approximates per-task learnability from a short probe run and two endpoint evaluations. \textsc{TrajVal} can be used either as a standalone static prior for task sampling or as a multiplicative prior for existing online schedulers. Experiments on mathematical and logical reasoning benchmarks across multiple model scales show that \textsc{TrajVal} improves data efficiency over uniform sampling and provides complementary gains when combined with online scheduling methods.
\end{abstract}

%% file: section/01_intro.tex
\section{Introduction}
\label{sec:intro}
Reinforcement learning (RL) has become a central post-training 
paradigm for eliciting reasoning capabilities in large language 
models~\citep{deepseek-r1, openai-o1, qwen3}. A recurring concern in this setting is data efficiency: training pools contain large numbers 
of tasks whose contributions to learning differ substantially, 
yet uniform sampling allocates compute without regard to this 
heterogeneity. A growing body of work on \emph{task-value 
estimation} addresses this by weighting or scheduling tasks 
during training to accelerate 
convergence~\citep{dapo, pcl, bots, greso, hive}.
These methods predominantly rely on the model's instantaneous performance as the primary indicator of task value. By monitoring metrics such as pass rates or rewards at the current training state, they seek to prioritize tasks based on their immediate difficulty or ``solvability''.

However, current solvability provides only a partial view of task value. Two tasks with similarly low pass rates at the current policy state can still differ sharply in how they respond to continued optimization. In this work, we study this residual axis as \emph{learnability}: a task's expected positive response to continued training under a fixed post-training regime. Importantly, we do not treat learnability as an immutable property of a task in isolation. Rather, it is a \emph{regime-conditional} quantity, defined with respect to a given model family, initialization, optimization algorithm, reward function, data source, and training budget.

This notion is complementary to \emph{solvability}. Solvability asks how well the current policy already performs on a task; learnability asks whether that task is likely to improve productively under additional training. The distinction matters operationally: among tasks with similar current solvability, some continue to yield positive reward response under training, while others remain flat, saturated, or even regress. Existing scheduling methods mainly target the former quantity, which can leave useful information about future training response untapped.

In this work, we study \textbf{learnability} as a complementary dimension of task value. By analyzing training traces, we observe that the policy exhibits highly heterogeneous learning responses to different tasks: while performance on some tasks reliably improves as training progresses, performance on others stagnates or degrades. Empirical analysis demonstrates that this divergence is both context-robust and predictive of RL post-training gains: each task's learnability reproduces across independent training runs sampled from the same data distribution, and yields measurable improvements when leveraged to select or weight tasks.

Directly measuring this signal is expensive: it requires tracking each task's performance trajectory over a full post-training run that sweeps the entire dataset for multiple epochs. To turn learnability into a prior usable before RL post-training begins, we apply two simplifications: summarizing the trajectory by a few performance values along it, and replacing the full post-training run with a short proxy run used to obtain those values. Building on these two reductions, we propose \textbf{\textsc{TrajVal}}, a preprocessing-stage estimator that yields a continuous sampling prior. It can serve as a standalone method for task weighting and composes multiplicatively with existing online task-scheduling methods.

We evaluate \textsc{TrajVal} on mathematical and logical reasoning benchmarks across multiple model scales. Empirical results show that it consistently outperforms uniform sampling and provides stable improvements over competitive task-value baselines when used as a plug-in prior. 
Our code is available at an anonymous repository: \href{https://github.com/SYSUzhouting/Task-Learnability}{https://github.com/SYSUzhouting/Task-Learnability}.

Our contributions are threefold:
\begin{itemize}[leftmargin=*]
    \item We distinguish \emph{learnability} from snapshot-based \emph{solvability} in RL post-training, and operationalize learnability as a task's regime-conditional positive response to continued training.
    \item Empirical evidence shows that \emph{learnability} is reproducible across independently sampled contexts and predictive of downstream utility, even among tasks with matched initial solvability.
    \item We propose \textsc{TrajVal}, a lightweight probe-based estimator of learnability that can be used either as a standalone static prior or as a multiplicative prior for existing online task schedulers.
\end{itemize}

%% file: section/02_related_works.tex
\section{Related Work}

\noindent\textbf{Static Task-Value Curation.} Offline task-value curation scores the data pool before training. In supervised fine-tuning, methods such as LIMA, LIMO, and s1 select high-value subsets using base-model pass rates or external model assessments~\citep{lima, limo, s1}. Similar ideas have been explored in RL using signals such as base-model uncertainty~\citep{zhaoufo}, complexity~\citep{scalingrl}, and difficulty tiers~\citep{e2h}. These approaches rely on a fixed snapshot of the initial model and therefore provide limited information about how a task will respond to continued optimization. LIMR~\citep{limr} moves toward trajectory-aware valuation, but it requires a full training pass over the pool.

\noindent\textbf{Dynamic Task-Value Scheduling.} Dynamic schedulers instead estimate task value online as training progresses. Representative approaches use rollout-based pass-rate signals or model-based approximations such as value models and Bayesian posteriors~\citep{dapo, dots, pcl, bots, mopps}, while others filter prompts using reward consistency or single-step policy improvement signals~\citep{greso, hive, actorcurator, sec}. These methods provide adaptive feedback from the current policy state, but mainly emphasize instantaneous task solvability.

\noindent\textbf{Learning Dynamics and Our Positioning.} The use of training dynamics to characterize samples has a long history in supervised learning. Dataset Cartography~\citep{swayamdipta2020dataset} and EL2N~\citep{paul2021deep}, for example, show that the evolution of loss or confidence under training reveals information about sample difficulty, learnability, and generalization impact that is not captured by single-state evaluation. Our work brings this trajectory-based perspective to RL post-training by focusing on a lightweight estimate of task learnability that complements existing state-based scheduling signals.

%% file: section/03_learnability.tex
\begin{figure*}[t]
  \centering
  \begin{minipage}[c]{0.44\linewidth}
    \centering
    \includegraphics[width=\linewidth]{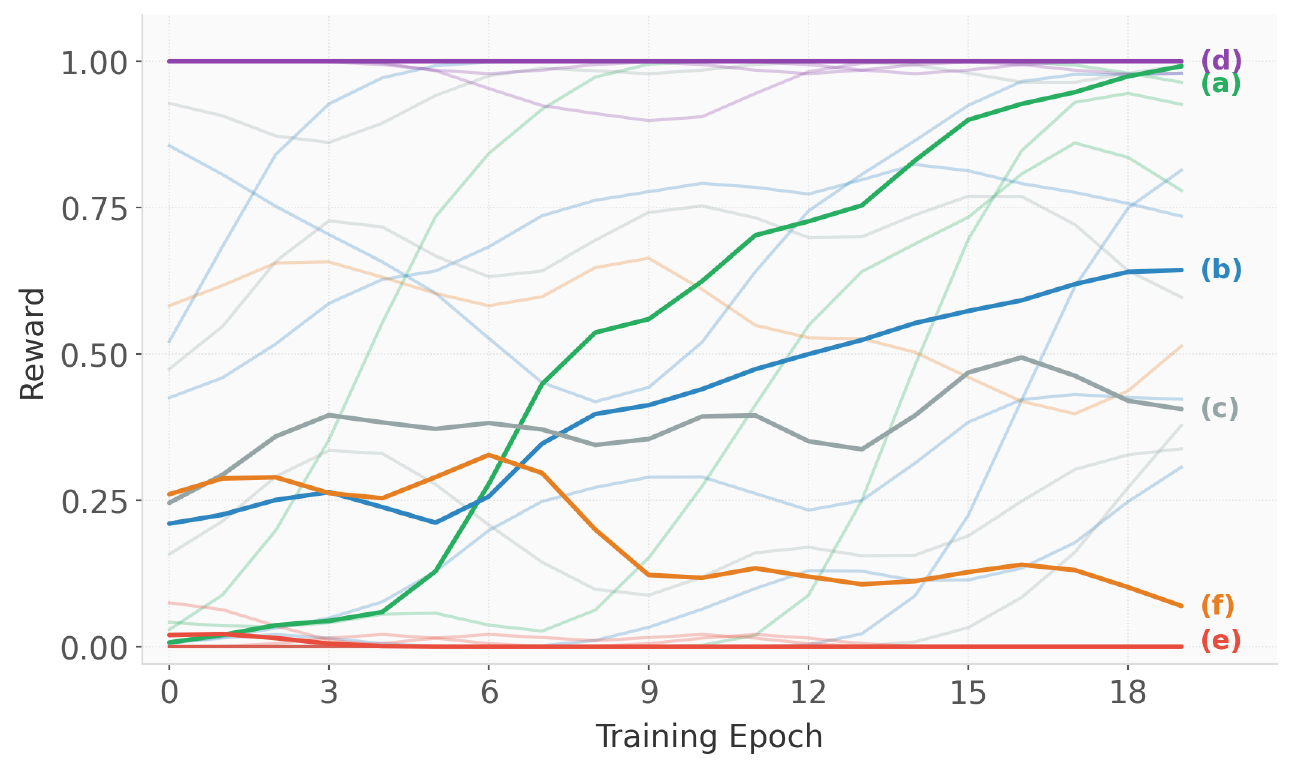}
  \end{minipage}
  \hfill
  \begin{minipage}[c]{0.55\linewidth}
    \centering
    \renewcommand{\arraystretch}{1.15}
    \resizebox{\linewidth}{!}{%
    \begin{tabular}{l l l c}
      \toprule
      \textbf{Profile} & \textbf{Criteria} &
      \textbf{Behavior} & \textbf{Learn.} \\
      \midrule
      (a) \textsc{Stable}
        & $\Delta{>}0.4$,\newline $R^{2}{>}0.4$ or $\sigma{\leq}0.35$
        & Steady rise
        & \checkmark \\
      (b) \textsc{Sluggish}
        & $\Delta{\in}(0.15, 0.4]$ or $\sigma{>}0.35$
        & Gradual rise
        & \checkmark \\
      (c) \textsc{Ineffective}
        & $\Delta{\in}({-}0.05, 0.15]$,\newline $\sigma{\leq}0.35$
        & Flat / noisy
        & \checkmark \\
      \midrule
      (d) \textsc{Mastered}
        & $\bar{r}_{\mathrm{early}}{>}0.9$, $\Delta{<}0.1$
        & Near-ceiling
        & \ding{55} \\
      (e) \textsc{Unlearned}
        & $\bar{r}_{\mathrm{late}}{<}0.1$, $\Delta{<}0.1$
        & Near-floor
        & \ding{55} \\
      (f) \textsc{Forgetting}
        & $\Delta{<}{-}0.05$
        & Net decline
        & \ding{55} \\
      \bottomrule
    \end{tabular}%
    }
  \end{minipage}

    \caption{%
      Representative per-task reward trajectories from the diagnostic pool $\mathcal{D}$ over 20 training epochs. Bold lines highlight six coarse trajectory patterns used for diagnosis. The right panel summarizes the corresponding profile descriptions. The profile distribution over $\mathcal{D}$ is shown in Figure~\ref{fig:stability_qwen3}(a).}
  \label{fig:traj_main}
\end{figure*}

\section{A Learnability-Based View of Task Value in RL Post-Training}
\label{sec:learnability}

In this section, we develop a learnability-based view of task value in two stages. We first examine how each task's reward evolves over the course of RL training, and show that these per-task trajectories motivate \emph{learnability} as a task-level signal distinct from snapshot-based solvability (\S\ref{sec:heterogeneity}); we then show that this signal is both \textbf{reproducible across independent training contexts drawn from the same data distribution} and \textbf{predictive of downstream utility} (\S\ref{sec:empirical_case}).

\subsection{Heterogeneous Per-Task Learnability under Training}

\label{sec:heterogeneity}
To examine per-task reward trajectories, we uniformly sample a diagnostic pool $\mathcal{D}$ of 2,048 mathematical reasoning tasks and train Qwen3-1.7B with GRPO for 20 epochs under a fixed training recipe (prompt batch size 64, 16 rollouts per prompt; full details in Appendix~\ref{appendix:exp_details}). As Figure~\ref{fig:traj_main} shows, tasks with similar initial pass rates can follow sharply different reward trajectories under continued training. This heterogeneity is temporal: it is not visible from a single snapshot of current pass rate alone.

This observation motivates our central distinction. A task's \emph{solvability} describes the current policy's performance on that task, whereas its \emph{learnability} describes how positively the task responds to continued optimization under a fixed regime. Intuitively, tasks that exhibit a steady, sustained upward trend with substantial reward growth exhibit high learnability, as they provide consistent and informative optimization signals. Conversely, tasks whose trajectories are highly volatile, stagnant, or declining exhibit low learnability, suggesting that the current optimization regime either fails to extract constructive signals from these tasks or struggles to consolidate the resulting updates.

To visualize this heterogeneity, we summarize each full reward trajectory using three simple descriptors---reward gain $\Delta$, standard deviation $\sigma$, and linear-fit $R^2$---and use them to define six coarse trajectory profiles (Figure~\ref{fig:traj_main}). These profiles are intended purely as a \textbf{diagnostic lens}: they help summarize representative trajectory shapes, but they are \emph{not} the formal definition of learnability, nor are they meant to form an exhaustive taxonomy of all possible training dynamics.

For descriptive analysis, we group profiles with non-negative training response (e.g., \textsc{Stable}, \textsc{Sluggish}, and \textsc{Ineffective}) separately from saturated, low-response, or regressive profiles (e.g., \textsc{Mastered}, \textsc{Unlearned}, and \textsc{Forgetting}). This binary grouping is used only for diagnosis in this section. In Section~\ref{sec:method}, we will replace these coarse categories with a continuous score used for actual data allocation.

\begin{figure*}[t]
    \centering
    \includegraphics[width=0.95\textwidth]{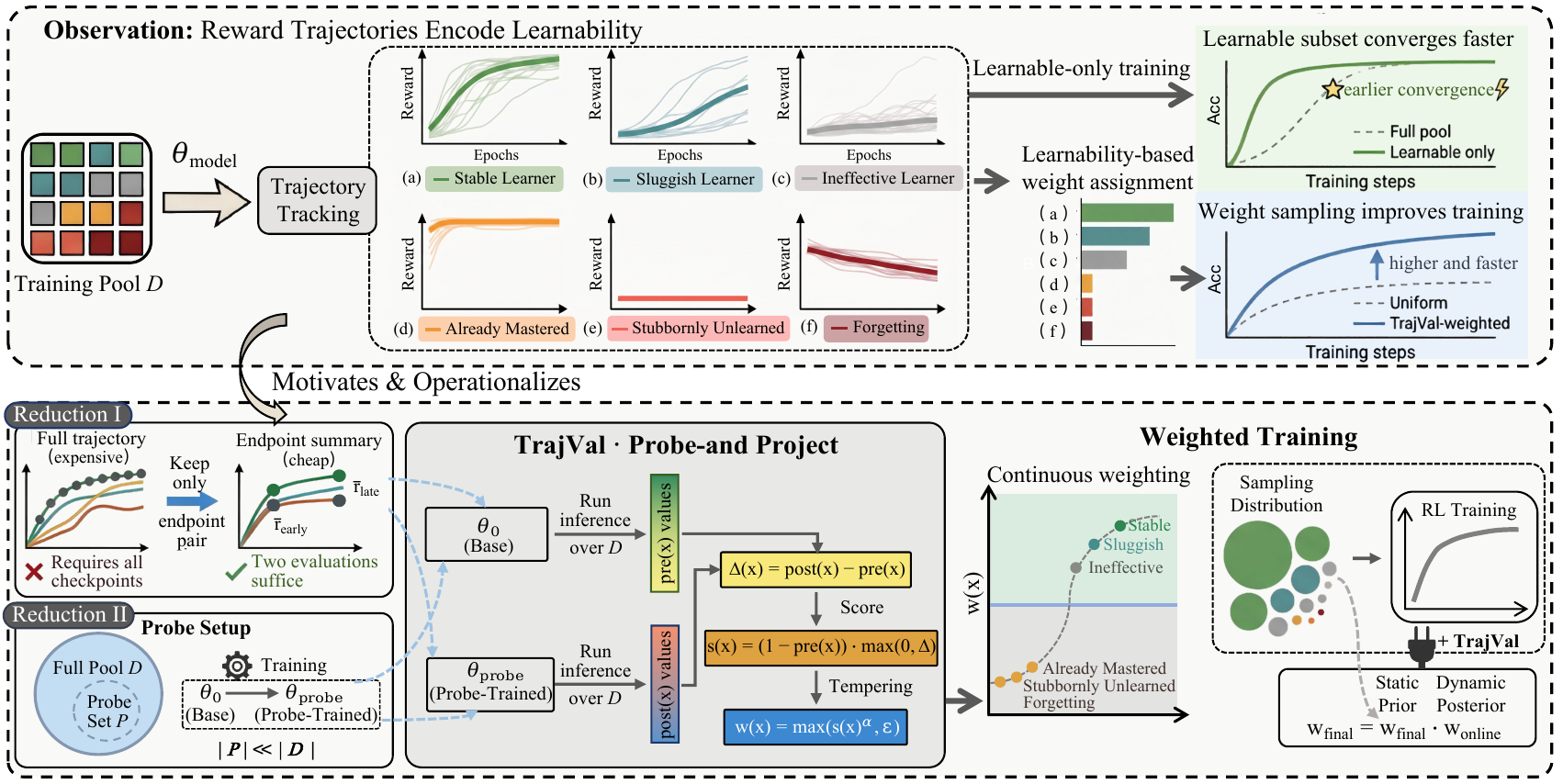}
    \caption{Overview of \textsc{TrajVal}. Top: per-task reward trajectories reveal heterogeneous training response beyond initial solvability. Bottom: \textsc{TrajVal} approximates this response using two reductions: from dense trajectories to an endpoint-based summary, and from full-pool training to a lightweight probe run. The resulting score is used as a static prior for prior-weighted sampling in RL training, either alone or in combination with an online scheduler.}
    \label{fig:overview}
\end{figure*}

\subsection{Learnability is Context-Robust and Predictive of 
Training Value}
\label{sec:empirical_case}
We next validate whether the learnability signal suggested by these trajectories satisfies two properties required for practical use: (\emph{i}) whether it is reproducible across independently sampled training contexts under a fixed regime, and (\emph{ii}) whether it predicts downstream utility for data allocation. We provide empirical evidence for both using Qwen3-1.7B with GRPO.

\paragraph{Cross-context reproducibility.}
We construct two probe sets $\mathcal{P}_1$ and $\mathcal{P}_2$ (each of size 512, uniformly sampled from $\mathcal{D}$, with 256 shared tasks) and independently train on $\mathcal{D}$, $\mathcal{P}_1$, and $\mathcal{P}_2$ for 20 epochs under a \emph{fixed training recipe} (same model, algorithm, and hyperparameters). Despite differences in mixture and the inherent stochasticity of RL, the macroscopic profile distributions remain nearly identical across the three contexts (Figure~\ref{fig:stability_qwen3}(a)--(c) in Appendix~\ref{app:stability_profiles}). At the individual level, the 256 shared tasks receive consistent profile assignments: Cohen's $\kappa$ reaches \textbf{0.776} across the six profiles and \textbf{0.879} under the binary learnable grouping (Figure~\ref{fig:stability_qwen3}(d)--(e)). The same pattern also appears under Llama-3.2-3B (Appendix~\ref{app:stability_family}), supporting the view that learnability is a \textbf{context-robust task-level signal} rather than an artifact of any particular mixture.

\begin{figure}[h]
    \centering
    \includegraphics[width=\linewidth]{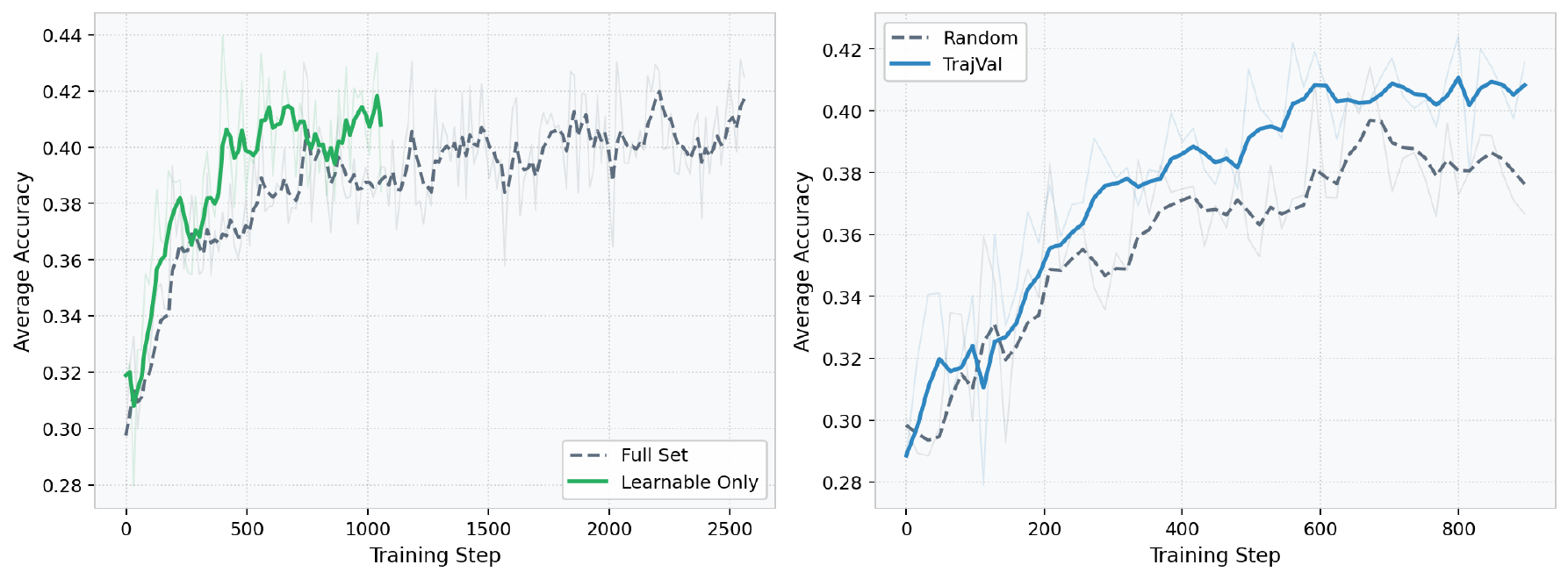}
    \caption{Effect of learnability-based data allocation on Qwen3-1.7B with GRPO, averaged over six mathematical reasoning benchmarks. \textbf{Left:} Training on the learnable subset versus the full pool over 20 epochs. \textbf{Right:} An oracle-based soft prior over the full pool versus random sampling at matched training budget.}
    \label{fig:sec32}
\end{figure}

\paragraph{Predictive value for downstream utility.}
% \qianli{The connection between the title and the evidence is not immediately clear. If the intended claim is that learnability predicts which tasks are more valuable for downstream RL training, I would also consider renaming the paragraph to something more direct, such as ``Learnability Predicts Downstream Training Value'' or ``Learnability Identifies High-Value Training Tasks.''}
We next ask whether the learnability grouping translates into downstream utility under two complementary operationalizations, measured by average accuracy across six mathematical reasoning benchmarks (details in Appendix~\ref{appendix:evaluation_and_metrics}).
\textit{As a hard cut}, restricting training to tasks marked \emph{learnable} ($\approx 41\%$ of $\mathcal{D}$) matches the full pool's converged accuracy ($\approx 0.41$) in \textbf{$3.6\times$ fewer steps} (Figure~\ref{fig:sec32}, left). \textit{As a soft prior over the full pool}, an oracle-based scheme that assigns higher sampling probability to learnable profiles---without discarding any task---outperforms random sampling at matched training budget after both have converged ($0.410$ vs $0.387$, Figure~\ref{fig:sec32}, right). The signal therefore yields measurable gains under both pruning-style and weighting-style uses, suggesting that its utility is not specific to either form of data allocation. In Section~\ref{sec:within_bin}, we further isolate learnability from initial solvability through a controlled analysis.

Together, these results suggest that learnability behaves as a reproducible and practically useful dimension of task value under a fixed training regime. The remaining challenge is computational: obtaining this signal at oracle quality requires a full $T$-epoch training pass over $\mathcal{D}$, which largely offsets the efficiency gains it can provide. Section~\ref{sec:method} therefore asks whether a much cheaper probe procedure can preserve the task ordering most relevant to learnability-aware data allocation.

%% file: section/04_method.tex
\section{\textsc{TrajVal}: A Probe-Based Estimator of Learnability}
\label{sec:method}

\begin{figure}[t]
    \centering
    \includegraphics[width=0.43\textwidth]{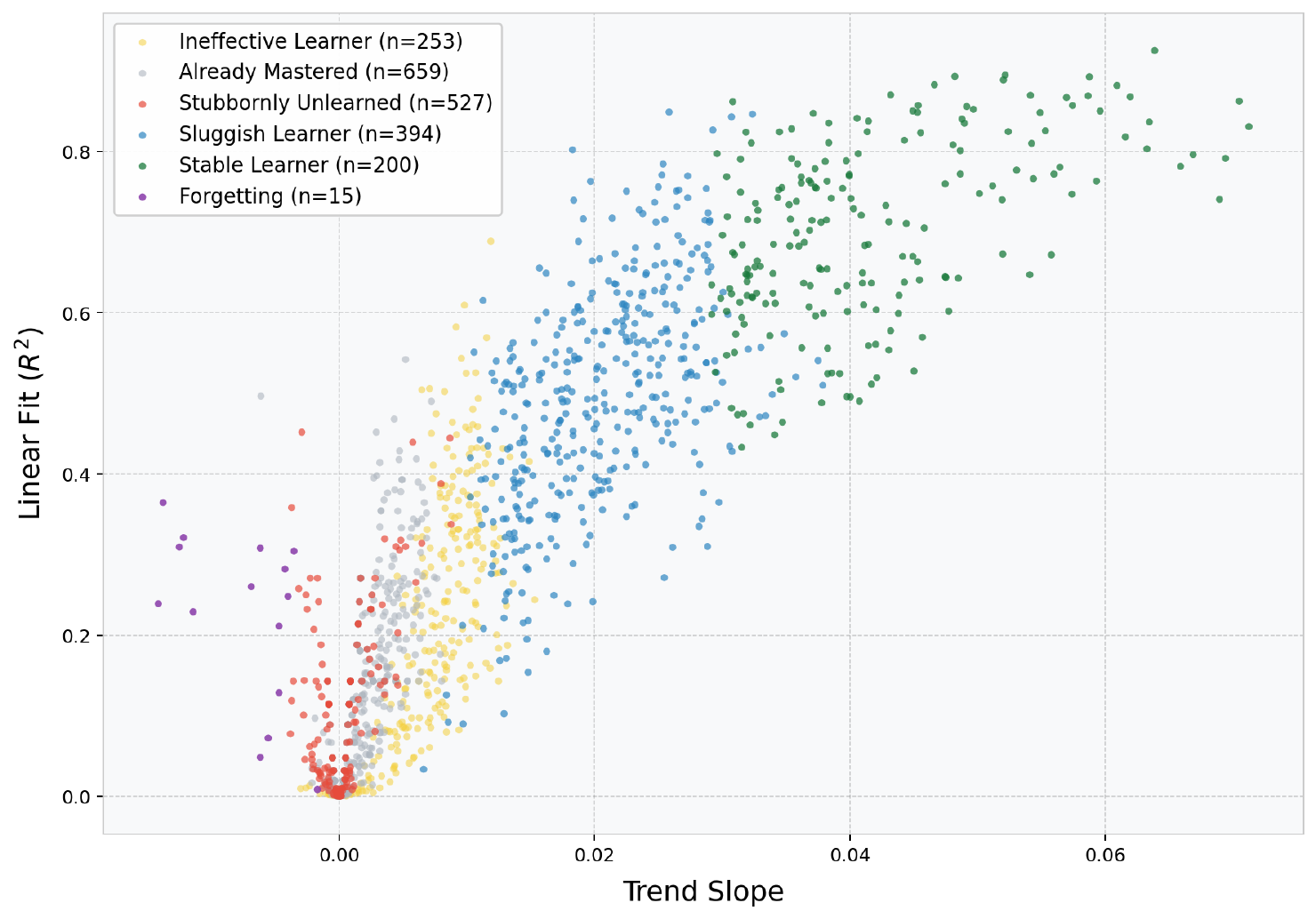}
    \caption{All tasks in $\mathcal{D}$ projected onto the slope--$R^{2}$ space of their training trajectories. Colors indicate profile labels derived from the endpoint pair $(\bar{r}_{\mathrm{early}}, \bar{r}_{\mathrm{late}})$.}
    \label{fig:slope_r2}
\end{figure}

Section~\ref{sec:learnability} used full reward trajectories to characterize an oracle notion of learnability. \textsc{TrajVal} does not redefine that notion; rather, it provides a lightweight \textbf{estimator} of it. The key question is therefore not whether a small number of evaluations can reconstruct every detail of the full trajectory, but whether they preserve the task ordering most relevant to learnability-aware data allocation.

To make oracle learnability practical before the main RL run begins, \textsc{TrajVal} applies two reductions. First, it replaces dense trajectory tracking with an endpoint-based summary of training response. Second, it replaces full-pool training with a much cheaper probe training run on a small uniformly sampled subset. Together, these reductions yield a static learnability prior that can be computed offline and used either on its own or in combination with existing online schedulers.

\subsection{Lightweight Estimation}
\label{sec:reduction}

\paragraph{Reduction I: From Trajectories to Endpoint-Based Summaries.}

Continuous checkpoint evaluation is expensive. We therefore ask whether the task ordering relevant to learnability can be captured by a much smaller summary: the mean rewards over an early phase and a late phase, denoted $\bar r_{\mathrm{early}}$ and $\bar r_{\mathrm{late}}$, together with their difference $\Delta = \bar r_{\mathrm{late}} - \bar r_{\mathrm{early}}$. This summary no longer preserves the full trajectory shape, but it still retains the minimal temporal contrast needed to distinguish a task's current solvability from how it responds over training. The goal is thus not to reconstruct the entire reward curve, but to preserve the distinctions most relevant to learnability-aware allocation.

Using this endpoint pair, we derive a simplified endpoint-only approximation to the diagnostic profiles in Figure~\ref{fig:traj_main}, bypassing statistics such as $\sigma$ and $R^2$ that require intermediate checkpoints. To test whether this summary retains the global structure of trajectory behavior, we project all tasks in $\mathcal{D}$ onto the slope--$R^2$ space computed from their full reward sequences, which is independent of the endpoint approximation. As Figure~\ref{fig:slope_r2} shows, labels derived solely from the endpoint pair remain well separated in this endpoint-independent space, indicating that the summary preserves the distinctions most relevant to learnability estimation.

\paragraph{Reduction II: From Full-Pool Training to Probe Training.}

Even with the endpoint summary, estimating $\bar r_{\text{late}}$ from full-pool training would still require a complete $T$-epoch pass over $\mathcal{D}$. We replace this with a much cheaper proxy. Let $\mathcal{P} \subset \mathcal{D}$ be a small uniformly sampled probe set with $|\mathcal{P}| \ll |\mathcal{D}|$. Because $\mathcal{P}$is drawn from the same training distribution, optimizing the base model $\theta_0$ on $\mathcal{P}$ exposes the policy to a representative subset of the optimization signals present in $\mathcal{D}$. The resulting probe-trained model $\theta_{\text{probe}}$ can therefore serve as a practical proxy for a later training state.

This substitution is empirical rather than axiomatic, so we verify it directly. For each task $x \in \mathcal{D}$, we compare probe-derived endpoint estimates against oracle values obtained from the full 2,048-task profiling run. The resulting Spearman correlations are $\rho = 0.940$ for $\bar r_{\text{early}}$ and $\rho = 0.876$ for $\bar r_{\text{late}}$, indicating that the relative ordering needed for learnability-aware allocation largely survives this substitution.

\subsection{Implementation: From Probe Estimators to Sampling Weights}
\label{sec:implementation}
Given probe-based estimates of the early and late rewards for every task $x \in \mathcal{D}$, we now convert them into a sampling distribution. 
Concretely, we replace the coarse profile-based diagnostic view from Section~\ref{sec:heterogeneity} with a continuous learnability score, as illustrated in Figure~\ref{fig:overview} and Algorithm~\ref{algo:main}.

\paragraph{Endpoint Estimation.}
For each task $x \in \mathcal{D}$, we summarize the probe trajectory by averaging per-task accuracies over $K$ checkpoints from an early window and a late window:
\begin{equation}
\small
\bar{r}_{\phi}(x) \;=\; \frac{1}{K}\!\!\sum_{t \in \mathcal{T}_{\phi}} \mathrm{acc}(x,\, \theta_t),
\quad \phi \in \{\mathrm{early},\,\mathrm{late}\}
\end{equation}

The net improvement is then captured by their difference, $\Delta(x) = \bar{r}_{\mathrm{late}}(x) - \bar{r}_{\mathrm{early}}(x)$. The window size $K$ is a preprocessing-budget hyperparameter that trades additional inference passes for reduced rollout variance. In our main experiments, we evaluate the most economical setting $K=1$; any gain observed there carries over to higher-$K$ settings with lower estimator noise.

\paragraph{Continuous Learnability Scoring.}
To convert the probe statistics into per-task sampling weights, we define a simple \emph{learnability-aware task-value proxy} that combines two desiderata from \S\ref{sec:heterogeneity}: tasks should still have room to improve, and they should exhibit a positive response under training:
\begin{equation}
\label{eq:trajval-score}
s(x) \;=\; \underbrace{\bigl(1 - \bar{r}_{\mathrm{early}}(x)\bigr)}_{\text{learning headroom}} \;\cdot\; \underbrace{\max\!\bigl(0,\;\Delta(x)\bigr)}_{\text{directional alignment}}.
\end{equation}
The directional term suppresses tasks whose reward remains flat or declines under probe training, while the headroom term down-weights tasks that are already near saturation. Their product therefore favors tasks that are both still improvable and already showing evidence of constructive response.

The role of the headroom term is not to replace $\Delta(x)$ as the indicator of improvement, but to encode an allocation preference: among tasks with comparable positive gain, higher priority is assigned to those with greater remaining room for improvement under the main training budget. Accordingly, the score places a modest penalty on near-mastered tasks whose short-term gains may be positive but whose residual training value is limited. Additional intuition for this score design is provided in Appendix~\ref{app:score_intuition}.

\paragraph{Sampling Integration and Diversity.}
Scores $s(x)$ can span a wide range, and direct proportional sampling risks over-concentrating on a narrow high-score subset. To retain learnability-aware priority while preserving distributional coverage, we apply power compression and a probability floor:
\begin{equation}
\small
w(x) \;=\; \max\!\bigl(s(x)^{\alpha},\;\epsilon\bigr), \quad
p(x) \;=\; \frac{w(x)}{\displaystyle\sum_{x' \in \mathcal{D}} w(x')}.
\end{equation}
The exponent $\alpha < 1$ compresses the score range, limiting the weight ratio between high- and low-scoring tasks. The floor $\epsilon$ ensures that every task retains a minimal sampling probability, including those with near-zero learnability scores---a deliberate choice that maintains exposure to the full data distribution and guards against degenerate concentration.

\paragraph{Integration as a Multiplicative Prior.}
\textsc{TrajVal} provides a static learnability prior that composes naturally with existing online schedulers. For a method with dynamic sampling weight $w_{\text{online}}(x,t)$, we use
\begin{equation}
w_{\text{final}}(x,t)=w_{\textsc{TrajVal}}(x)\cdot w_{\text{online}}(x,t).
\end{equation}
This combination is complementary in spirit: \textsc{TrajVal} supplies a stable offline prior from initialization, while the online scheduler adapts to the evolving policy state during training. \textsc{TrajVal} therefore targets a failure mode that online methods do not directly address, namely poor allocation before sufficient online evidence has accumulated.

%% file: section/05_exp.tex
\section{Experiments}
\label{sec:experiments}
\subsection{Experimental Setup}
\label{sec:setup}

We validate \textsc{TrajVal} across two domains, three model variants spanning different scales and families, and multiple evaluation benchmarks. Full implementation details are provided in Appendix~\ref{appendix:exp_details}.

\textbf{Tasks and Datasets.}
We evaluate across two distinct domains: \textbf{mathematical reasoning}, which requires rigorous multi-step numerical deduction, and \textbf{logical reasoning}, which demands structured, abstract inference over diverse puzzle formats. We train on \textbf{DAPO-Math}~\citep{dapo} (17,398 tasks) for math, and the \textbf{Logic subset} (4,998 tasks) of the GURU~\citep{guru} dataset for logic, which spans ordering puzzles, graph puzzles, ARC-AGI~\citep{arc-1,arc-2}, and BARC~\citep{barc} tasks, with exact-match binary reward.

\textbf{Evaluation and Metrics.}
For \textbf{math}, we assess on six benchmarks: AIME24, AIME25, AMC23, MATH500~\citep{math500}, Minerva~\citep{minerva}, and OlympiadBench~\citep{olympiadbench}. For \textbf{logic}, we evaluate on a held-out set from Ordering Puzzles and ARC-AGI~\citep{arc-1}. We report three complementary metrics per benchmark, as well as their macro-average (\textbf{Avg}): (1)~\textbf{Best Acc}, the peak sliding-window mean accuracy; (2)~\textbf{AUC}, the time-averaged accuracy across all training steps; and (3)~\textbf{Steps-to-Baseline (S2B)}, the fraction of steps required for a \textsc{TrajVal}-augmented method to first match the Best Acc of its corresponding baseline (lower is better). All results are means over 2 seeds.

\textbf{Implementation.}
All experiments are built on the Trinity-RFT~\citep{trinity} framework with \textbf{GRPO}~\citep{grpo}. Our main results use \textbf{Qwen3-1.7B} and \textbf{Qwen3-4B}~\citep{qwen3}, while \textbf{Llama-3.2-3B}~\citep{llama3} is included as an additional cross-family validation in Appendix~\ref{app:stability_family}. In both domains, RL is applied with $n=16$ rollouts per prompt. For \textsc{TrajVal}, the learnability prior is derived from two inference passes over the base model $\theta_0$ and a lightweight probe-trained model $\theta_{\text{probe}}$; the sampling weight is parameterized by a power-compression exponent $\alpha=0.3$ and a smoothing floor $\epsilon=0.05$.

\input{section/Tables/exp_main_both}   

\textbf{Baselines.}
We compare against three representative task-selection strategies: (1)~\textbf{\textsc{Base-GRPO}}~\citep{grpo}, vanilla GRPO with uniform task sampling; (2)~\textbf{\textsc{BOTS}}~\citep{bots}, a Bayesian online scheduler that maintains posterior estimates of per-task difficulty; and (3)~\textbf{\textsc{GRESO}}~\citep{greso}, a pre-rollout filtering method that skips temporally uninformative prompts. Baseline-specific settings follow the original works and are detailed in Appendix~\ref{appendix:baselines}.

%=============================================================

\subsection{Main Results}
\label{sec:main}
We evaluate \textsc{TrajVal} across two reasoning domains and three base training strategies. Table~\ref{tab:main_result} summarizes the main results, and Figure~\ref{fig:eval_curves} shows the corresponding evaluation curves. The overall pattern supports the central claim of the paper: task allocation for RL post-training benefits from modeling not only how solvable a task currently is, but also how positively it is likely to respond to continued optimization. Per-benchmark results are provided in Appendix~\ref{appendix:full_results}, and Appendix~\ref{app:rollout_dynamics} gives rollout-level views of training dynamics.

\textbf{TrajVal as a standalone method.}
As a direct replacement for uniform sampling, \textsc{TrajVal} consistently improves both convergence speed and peak performance over \textsc{Base-GRPO} across all four domain--scale configurations. The gains are visible in both Math and Logic, with especially large acceleration on Logic at 4B, where the method reaches the baseline peak in only $40\%$ of the steps.

\textbf{TrajVal as a plug-in prior.}
\textsc{TrajVal} also yields consistent gains when composed with existing online schedulers. Augmenting \textsc{BOTS} and \textsc{GRESO} improves Best Acc and AUC in every configuration, while also reducing S2B in all cases. This pattern supports the view that learnability provides a stable prior that complements online solvability estimates, particularly when those estimates are still noisy early in training.

\textbf{Consistency across domains and scales.}
The improvements of \textsc{TrajVal} persist across both reasoning domains and model scales. We further verify the same qualitative trend on an additional model family, Llama-3.2-3B, with supporting cross-family evidence in Appendix~\ref{app:stability_family}. We next provide a controlled analysis to directly test the central claim that this gain cannot be reduced to current solvability alone.

\begin{figure}[t]
    \centering
    \includegraphics[width=0.48\linewidth]{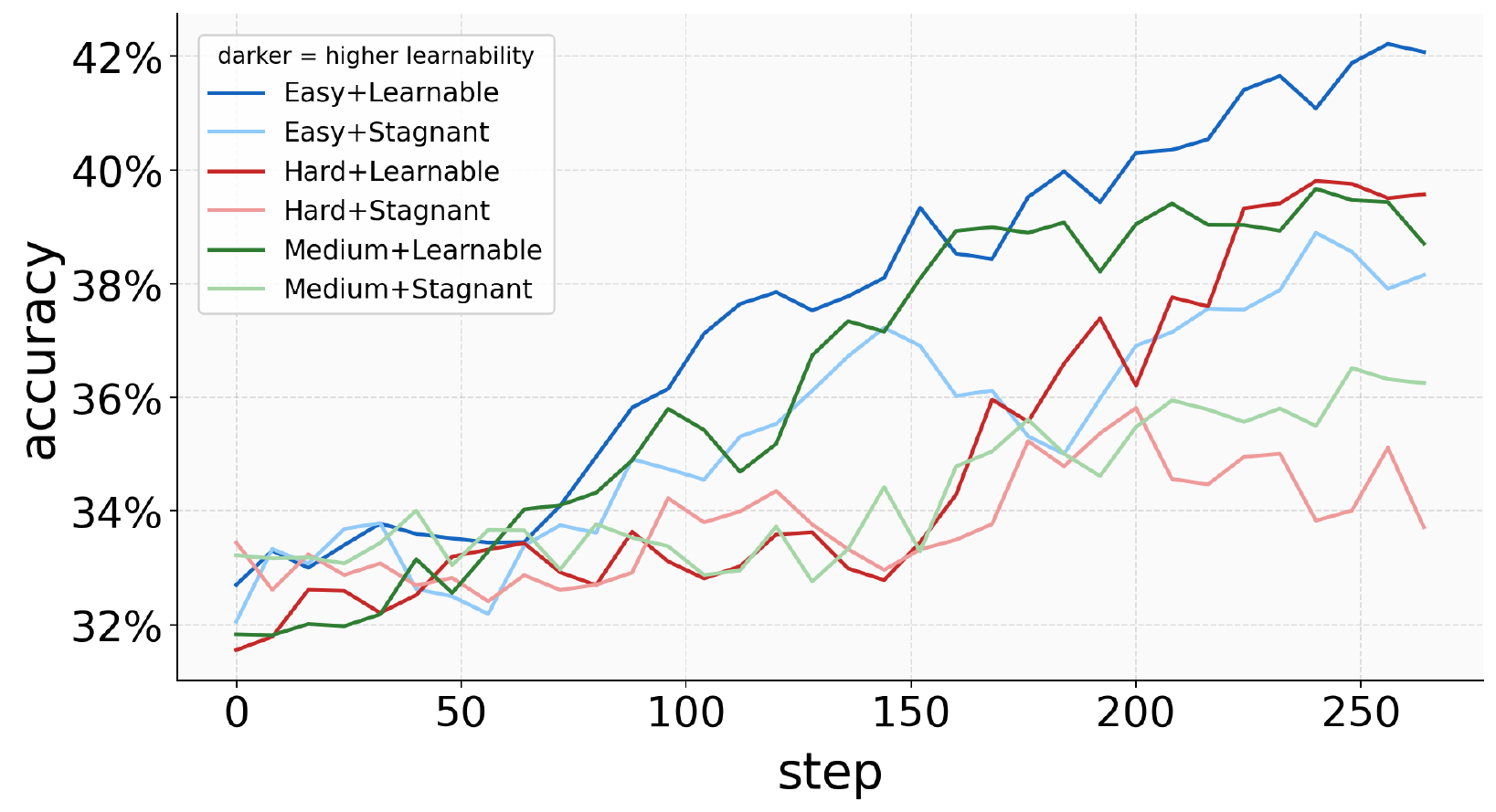}
    \hfill
    \includegraphics[width=0.48\linewidth]{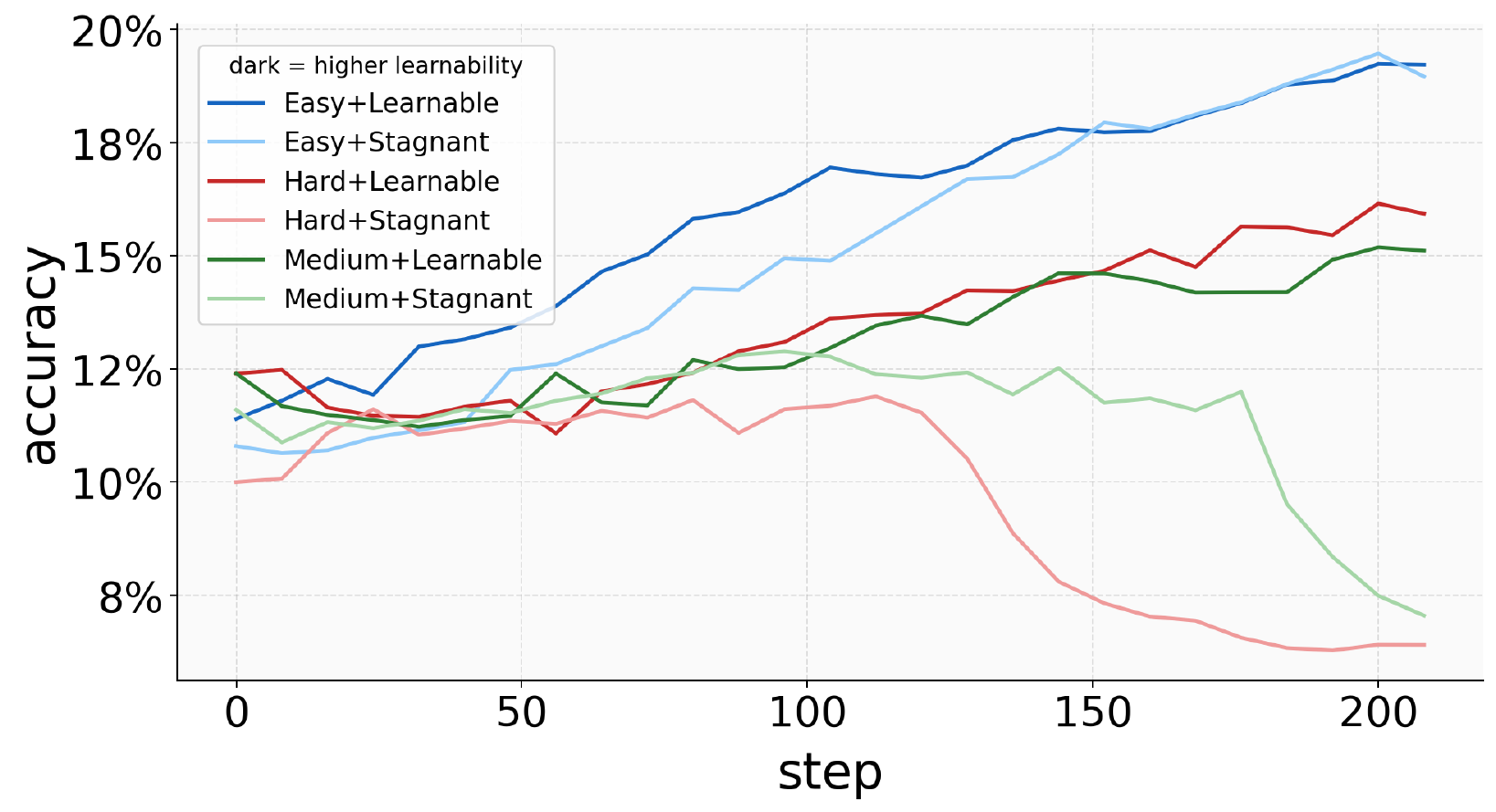}
    \caption{Within-pre-bin evaluation accuracy curves on Qwen3-1.7B with GRPO for high-$\Delta$ and low-$\Delta$ sub-pools: Math (left) and Logic (right).}
    \label{fig:within_bin_curves}
\end{figure}

\subsection{Within-Pre-Bin Analysis: Learnability Beyond Initial Solvability}
\label{sec:within_bin}

A central claim of this work is that learnability is not reducible to current solvability. We test this with a controlled within-bin analysis that isolates the effect of $\Delta(x)$ after stratifying by $\bar r_{\text{early}}(x)$. If high-$\Delta$ tasks still yield better downstream outcomes than low-$\Delta$ tasks within the same pre-bin, then the predictive signal cannot be explained by initial solvability alone.

\textbf{Setup.}
Within each domain, we partition the training pool into \emph{3 $\mathrm{pre}$-bins}: low ($\mathrm{pre} \in [0, 1/3]$), mid ($\mathrm{pre} \in (1/3, 2/3]$), and high ($\mathrm{pre} \in (2/3, 1]$). Within each bin, tasks are split by $\Delta(x)$ into two equal-size sub-pools---\textbf{high-$\Delta$} (top 50\%) and \textbf{low-$\Delta$} (bottom 50\%)---with average $\mathrm{pre}$ matched between the two, so that the only systematic difference is in $\Delta$. This controlled setup is intended to isolate the signal quality of $\Delta(x)$, rather than to replicate \textsc{TrajVal}'s weighted sampling scheme.

\input{section/Tables/exp_with_per_bin}

\textbf{Results.}
Table~\ref{tab:within_bin} and Figure~\ref{fig:within_bin_curves} show that across all six bin--domain combinations, \textbf{high-$\Delta$ sub-pools uniformly outperform their low-$\Delta$ counterparts} in both Best Acc and AUC, despite matched average $\mathrm{pre}$ within each bin. The effect is strongest in the Low-$\mathrm{pre}$ Math bin, where high-$\Delta$ tasks reach $0.434$ Best Acc versus $0.338$ for low-$\Delta$ tasks. On Math, meaningful gaps persist in the Mid and High bins; on Logic, the same monotonic ordering holds across all terciles. This shows that tasks with the same current solvability can still differ substantially in training value, and that this residual difference is captured by learnability.

\subsection{Ablation and Practical Analysis}
\label{sec:ablation}

We further examine \textsc{TrajVal} along five complementary dimensions: 

\begin{table}
\centering
\setlength{\tabcolsep}{4pt}
\resizebox{0.46\textwidth}{!}{%
\begin{tabular}{lccc|ccc}
\toprule
& \multicolumn{3}{c}{Math (Avg)} & \multicolumn{3}{c}{Logic} \\
\cmidrule(lr){2-4}\cmidrule(lr){5-7}
\textbf{Method} & S2B $\downarrow$ & Best Acc $\uparrow$ & AUC $\uparrow$ & S2B $\downarrow$ & Best Acc $\uparrow$ & AUC $\uparrow$ \\
\midrule
Random & -- & 0.3967 & 0.3620 & -- & 0.1826 & 0.1544 \\
\midrule
\multicolumn{7}{l}{\textit{Score composition}} \\
pre-only         & 85.86  & 0.4084 & 0.3675 & 79.08 & 0.1904 & 0.1625 \\
post-only        & 85.86  & 0.4067 & 0.3662 & 82.61 & 0.1897 & 0.1622 \\
$\Delta$-only    & 85.86  & 0.4086 & 0.3660 & 78.26 & 0.1864 & 0.1641 \\
\midrule
\multicolumn{7}{l}{\textit{Probe set size}} \\
$|\mathcal{P}|=256$  & 100.00 & 0.3968 & 0.3623 & \textbf{72.73} & \textbf{0.1916} & 0.1639 \\
$|\mathcal{P}|=1024$ & 90.91  & 0.4021 & 0.3670 & 73.91 & 0.1892 & 0.1638 \\
\midrule
\textbf{TrajVal} & \textbf{65.66} & \textbf{0.4105} & \textbf{0.3730} & 73.91 & 0.1904 & \textbf{0.1663} \\
\bottomrule
\end{tabular}}
\caption{Ablation results on Qwen3-1.7B over score composition and probe set size; $|\mathcal{P}|=512$ is the default.}
\label{tab:ablation_combined}
\end{table}

\textbf{Score composition.}
All three single-dimension variants---\textbf{pre-only}, \textbf{post-only}, and \textbf{$\Delta$-only}---underperform full \textsc{TrajVal}, especially in convergence speed on Math. The results in Table \ref{tab:ablation_combined} indicate that headroom and directional alignment are complementary: $\mathrm{pre}$ captures remaining room for improvement, while $\Delta$ captures responsiveness to optimization. Only their combination prioritizes tasks that are both improvable and productively learnable under the current training regime.

\textbf{Diagnostic analyses.}
Appendix~\ref{appendix:diagnostic} provides two additional findings. First, $s(x)$ concentrates sampling weight on learnable tasks ($67\%$ of total weight for $41\%$ of the pool), and the score distribution remains nearly unchanged across independent probe sets. Second, the score ranking stabilizes early: the Spearman correlation with the final ranking reaches $\rho \geq 0.85$ by probe epoch $k{=}14$ out of $T{=}20$ (Figure~\ref{fig:probe_epoch_stability}), suggesting the probe horizon can be shortened without changing score quality.

\textbf{Hyperparameter sensitivity.}
We examine two key hyperparameters. A moderate probe set size ($|\mathcal{P}|{=}512$, roughly $3\%$ of the pool) in Table \ref{tab:ablation_combined} provides the best overall trade-off. Across $\alpha \in \{0.1, 0.3, 0.5\}$, all settings improve over \textsc{Random} (Appendix~\ref{appendix:hyperparam}), indicating that \textsc{TrajVal} is not highly sensitive to the exact choice of $\alpha$.

\textbf{Computational overhead.}
\textsc{TrajVal} adds a one-time probe run on ${\approx}3\%$ of the pool and two inference passes for endpoint estimation, while the per-step sampling overhead is negligible (${\approx}0.01\%$ of total pipeline time; Appendix~\ref{appendix:compute_cost}). This marginal cost is far offset by the S2B reductions.

\textbf{Cross-regime reuse.}
We further examine whether a \textsc{TrajVal} prior estimated under one regime can be reused across model scales or families, rather than re-estimated for each target model (Appendix~\ref{sec:transfer}). Transfer retains useful signal, particularly within the same family, while in-regime estimation achieves the strongest overall performance.

Taken together, these results show that \textsc{TrajVal} is well-behaved in practice: its gains rely on the joint use of headroom and directional response, its ranking remains stable under lightweight probing, its performance is robust to hyperparameter variation, and its one-time preprocessing cost is offset by substantial reductions in steps to baseline and further amortized through cross-model prior reuse.

%% file: section/Tables/exp_main_both.tex
\begin{table*}[t]
\centering
\resizebox{\linewidth}{!}{%
\setlength{\tabcolsep}{5pt}
\begin{tabular}{l ccc ccc|ccc ccc}
\toprule
& \multicolumn{6}{c}{\textbf{Math}} & \multicolumn{6}{c}{\textbf{Logic}} \\
\cmidrule(lr){2-7}\cmidrule(lr){8-13}
& \multicolumn{3}{c}{Qwen3-1.7B} & \multicolumn{3}{c}{Qwen3-4B}
& \multicolumn{3}{c}{Qwen3-1.7B} & \multicolumn{3}{c}{Qwen3-4B} \\
\cmidrule(lr){2-4}\cmidrule(lr){5-7}\cmidrule(lr){8-10}\cmidrule(lr){11-13}
\textbf{Method}
& S2B $\downarrow$ & Best Acc $\uparrow$ & AUC $\uparrow$
& S2B $\downarrow$ & Best Acc $\uparrow$ & AUC $\uparrow$
& S2B $\downarrow$ & Best Acc $\uparrow$ & AUC $\uparrow$
& S2B $\downarrow$ & Best Acc $\uparrow$ & AUC $\uparrow$ \\
\midrule
Random (\textsc{Base-GRPO})
& --     & 0.3967 & 0.3620
& --     & 0.5060 & 0.4829
& --     & 0.1826 & 0.1544
& --     & 0.2052 & 0.1932 \\
Random+\textsc{TrajVal}
& 65.66  & \textbf{0.4105} & \textbf{0.3730}
& 60.00  & \textbf{0.5190} & \textbf{0.4931}
& 73.91  & \textbf{0.1904} & \textbf{0.1663}
& 40.00  & \textbf{0.2260} & \textbf{0.2101} \\
\midrule
BOTS~\cite{bots}
& --     & 0.4042 & 0.3648
& --     & 0.5076 & 0.4767
& --     & 0.1862 & 0.1556
& --     & 0.2216 & 0.2076 \\
BOTS+\textsc{TrajVal}
& 95.00  & \textbf{0.4081} & \textbf{0.3703}
& 75.00  & \textbf{0.5178} & \textbf{0.4863}
& 95.65  & \textbf{0.1866} & \textbf{0.1645}
& 60.87  & \textbf{0.2232} & \textbf{0.2112} \\
\midrule
GRESO~\cite{greso}
& --     & 0.4028 & 0.3621
& --     & 0.5062 & 0.4695
& --     & 0.1878 & 0.1568
& --     & 0.2244 & 0.2082 \\
GRESO+\textsc{TrajVal}
& 90.00  & \textbf{0.4083} & \textbf{0.3681}
& 75.00  & \textbf{0.5170} & \textbf{0.4794}
& 86.96  & \textbf{0.1914} & \textbf{0.1648}
& 78.26  & \textbf{0.2252} & \textbf{0.2103} \\
\bottomrule
\end{tabular}}
\caption{Main results across Math and Logic domains at two model scales. \textsc{TrajVal} is used either as a standalone static prior over uniform sampling or as a multiplicative prior for online schedulers.}
\label{tab:main_result}
\end{table*}

%% file: section/Tables/exp_with_per_bin.tex
\begin{table*}[t]
\centering
\resizebox{0.75\linewidth}{!}{%
\setlength{\tabcolsep}{5pt}
\begin{tabular}{ll cccc cccc}
\toprule
& & \multicolumn{4}{c}{\textbf{Math}} & \multicolumn{4}{c}{\textbf{Logic}} \\
\cmidrule(lr){3-6}\cmidrule(lr){7-10}
\textbf{Tercile} & \textbf{Sub-pool} & Avg $\mathrm{pre}$ & Avg $\Delta$ & Best Acc $\uparrow$ & AUC $\uparrow$ & Avg $\mathrm{pre}$ & Avg $\Delta$ & Best Acc $\uparrow$ & AUC $\uparrow$ \\
\midrule
\multirow{2}{*}{Low}
  & low-$\Delta$  & 0.0015 & 0.0140 & 0.338          & 0.3379          & 0.0000 & 0.0026 & 0.124          & 0.0946 \\
  & high-$\Delta$ & 0.0016 & 0.1080 & \textbf{0.434} & \textbf{0.3545} & 0.0000 & 0.0818 & \textbf{0.152} & \textbf{0.1367} \\
\midrule
\multirow{2}{*}{Mid}
  & low-$\Delta$  & 0.1660 & 0.0570 & 0.361          & 0.3449          & 0.0065 & 0.0401 & 0.134          & 0.1114 \\
  & high-$\Delta$ & 0.1660 & 0.2860 & \textbf{0.391} & \textbf{0.3675} & 0.0065 & 0.1384 & \textbf{0.150} & \textbf{0.1325} \\
\midrule
\multirow{2}{*}{High}
  & low-$\Delta$  & 0.6810 & 0.0480 & 0.387          & 0.3586          & 0.4695 & 0.2994 & 0.194          & 0.1550 \\
  & high-$\Delta$ & 0.6820 & 0.2000 & \textbf{0.398} & \textbf{0.3813} & 0.4693 & 0.4578 & \textbf{0.198} & \textbf{0.1635} \\
\bottomrule
\end{tabular}}
\caption{Within-pre-bin analysis on Qwen3-1.7B across Math and Logic. Matched Avg $\mathrm{pre}$ within each tercile isolates the effect of $\Delta$ on downstream performance.}
\label{tab:within_bin}
\end{table*}

%% file: section/06_conclusion.tex
\section{Conclusion}

We introduced learnability as a regime-conditional dimension of task value in RL that complements current solvability, and showed that tasks with similar initial pass rates can differ markedly in their response to continued optimization. To make this signal practical, we proposed \textsc{TrajVal}, a lightweight probe-based estimator that converts early/late endpoint statistics from a short probe run into sampling weights for main RL training. Across different tasks, models, and utilized settings, \textsc{TrajVal} improves data efficiency over uniform sampling and adds gains when combined with online schedulers. More broadly, our results suggest that task valuation for RL post-training should consider not only how solvable a task is now, but also how productively it is likely to improve under further training.

\section*{Limitations}
\textsc{TrajVal} is developed and evaluated on mathematical and logical reasoning tasks with binary reward signals; whether the learnability signal transfers equally well to domains with non-binary or model-graded rewards (e.g., open-ended generation evaluated by an LLM judge) remains an open question. In addition, our evaluation is confined to text-only settings, and extending \textsc{TrajVal} to multimodal RL post-training pipelines, where the method is in principle applicable, has not yet been explored. Finally, due to computational resource constraints, our experiments are conducted on models up to 8B parameters; validating the approach at larger model scales is left to future work.

\section*{Ethical Considerations}
This work focuses on improving the data efficiency of reinforcement learning post-training for large language models through learnability-based task scheduling. All base models used in our experiments (Qwen3-1.7B, Qwen3-4B, and Llama-3.2-3B) are publicly released under their respective open-source licenses, and all datasets employed (DAPO-Math and GURU) are publicly available research resources. We have used these resources in compliance with their stated terms. Generative AI tools were used only for limited language polishing and drafting assistance, while all technical content, experiments, analysis, and conclusions were produced and verified by the authors. Our research is methodological in nature and does not involve human subjects, personal data, or sensitive content. We do not foresee direct ethical risks arising from this work.

%% file: section/Appendix/implementation.tex
% \section{Implementation Details}

\section{Details of Experimental Setup}
\label{appendix:exp_details}

\subsection{Experimental Tasks and Datasets}
\label{appendix:tasks_and_data}

We validate our proposed method across two distinct domains to demonstrate its versatility: mathematical reasoning and logical reasoning. The former assesses the model's capacity for rigorous, multi-step numerical computation, while the latter evaluates its ability to perform structured, abstract reasoning across diverse puzzle types.

\paragraph{Mathematics}
For mathematical reasoning, the objective is to enhance the model's problem-solving capabilities. We use the \textbf{DAPO-Math-17k} dataset~\citep{dapo} for training. This dataset consists of 17,398 problems, primarily from competitive mathematics, with integer-based answers. The verified and binary reward structure (1.0 for a correct answer, 0.0 otherwise) provides a clear, objective signal for improving logical and computational accuracy.

\paragraph{Logic}
For logical and abstract reasoning, the objective is to improve the model's capacity for structured inference and pattern recognition across diverse reasoning formats. Training data is sourced from the \textbf{GURU}~\citep{guru} benchmark collection, a well-curated, cross-domain RL dataset. Specifically, we use the \textbf{Logic subset} (approximately 5.0k problems), which spans several complementary reasoning tasks: \textbf{Ordering Puzzles} (relational ordering under constraints), \textbf{Graph Puzzles} (implicit graph traversal and path reasoning), the public training splits of \textbf{ARC-AGI}~\citep{arc-1} and \textbf{ARC-AGI-2}~\citep{arc-2} (abstract grid transformation tasks), and a 3.4k-sample subset from \textbf{BARC}~\citep{barc} (synthetic ARC-style tasks for data augmentation). For all logic tasks, predictions are extracted from \texttt{<answer>} tags and evaluated via exact match, yielding a binary reward of 1.0 for a correct prediction and 0.0 otherwise.

\subsection{Evaluation and Metrics}
\label{appendix:evaluation_and_metrics}

\subsubsection{Evaluation Datasets.}

\paragraph{Mathematics.}
We evaluate mathematical reasoning on a suite of benchmarks commonly used for assessing Qwen-Math series~\citep{qwen2math} models. This selection covers a wide spectrum of difficulty, from challenging high-school problems to graduate-level and Olympiad-level questions:
\begin{itemize}[leftmargin=*]
    \item \textbf{AIME24} / \textbf{AIME25}: The American Invitational Mathematics Examination (AIME) features highly challenging, competition-level problems for high-school students.

    \item \textbf{AMC23}: A benchmark consisting of 40 problems from the American Mathematics Competitions (AMC). This widespread contest series for high-school students serves as a key test of foundational mathematical reasoning.
    
    \item \textbf{MATH500}~\citep{math500}: A classic and challenging benchmark comprising 500 problems from high-school math competitions, designed to test multi-step reasoning across algebra, combinatorics, geometry, number theory.

    \item \textbf{Minerva}~\citep{minerva}: A benchmark featuring 272 advanced problems from university and graduate-level STEM courses, covering subjects like differential equations and special relativity.

    \item \textbf{OlympiadBench}~\citep{olympiadbench}: An Olympiad-level benchmark containing 675 exceptionally difficult problems in mathematics and science that demand profound and complex reasoning abilities.
\end{itemize}

\paragraph{Logic}
We evaluate logical and abstract reasoning on a held-out evaluation set of 500 problems, drawn from two complementary benchmarks that probe distinct reasoning faculties:
\begin{itemize}[leftmargin=*]
    \item \textbf{Ordering Puzzles}: A set of relational reasoning problems that require the model to infer and satisfy ordering constraints among entities, serving as a probe for general structured logical inference.

    \item \textbf{ARC-AGI}~\citep{arc-1}: A benchmark of grid-based abstract reasoning tasks requiring the model to identify latent transformation rules from input--output examples and apply them to novel grids.
\end{itemize}

\subsubsection{Evaluation Metrics.}

All reported scores are averaged over 2 independent random seeds. A sampling temperature of $T=1.0$ is used throughout for generation~\citep{li2025can,guru}. To ensure robust estimates on stochastic benchmarks, we generate multiple solutions per problem for AIME 24/25 (8 generations) and AMC23 (4 generations). Online evaluation is performed every 12 steps for Mathematics and every 8 steps for Logic. Training curves are visualized with an Exponential Moving Average (EMA, factor 0.6) applied solely for display purposes; all quantitative metrics are computed on raw trajectories.

We report three complementary metrics for each benchmark as well as their macro-average across all benchmarks (\textbf{Avg}):

\begin{itemize}[leftmargin=*]
    \item \textbf{Best Acc}: The peak accuracy achieved over the full training run, computed as the maximum value of a sliding-window mean with a window of 5 consecutive evaluation steps applied to the raw accuracy curve. This windowed estimator reduces sensitivity to stochastic fluctuations at individual checkpoints. For the \textbf{Avg} column in Mathematics, Best Acc is the peak of the macro-averaged accuracy curve, computed by first averaging the seed-averaged per-benchmark trajectories across all benchmarks at each step and then applying the same sliding-window procedure.

    \item \textbf{AUC}: The time-averaged accuracy across the entire training run, computed as the arithmetic mean of all evaluation-step accuracies on the seed-averaged trajectory. This metric captures the overall training efficiency of a method, rewarding approaches that not only converge to high accuracy but do so consistently throughout training, rather than merely at the final checkpoint.

    \item \textbf{Steps-to-Baseline (S2B)}: A data-efficiency metric defined pairwise for each \textsc{TrajVal}-augmented method relative to its corresponding baseline. Let $T$ denote the Best Acc of the baseline and $s_{\text{base}}$ the training step at which the baseline's sliding-window mean first attains $T$. S2B is then the first step $s_{\text{comp}}$ at which the augmented method's sliding-window mean reaches $T$, expressed as a percentage of $s_{\text{base}}$:
    \begin{equation*}
        \text{S2B} = \frac{s_{\text{comp}}}{s_{\text{base}}} \times 100\%.
    \end{equation*}
    Lower values indicate that the augmented method matches the baseline's best performance with fewer training steps. For the \textbf{Avg} column in Mathematics, S2B is computed on the macro-averaged curve across all benchmarks.
\end{itemize}

\subsection{Training Details}
\label{appendix:train_details}

\paragraph{Framework and Infrastructure.}
All experiments are conducted on the \textbf{Trinity-RFT}~\citep{trinity} framework, utilizing nodes equipped with 8 NVIDIA A100 (80GB) GPUs. We employ PyTorch's Fully Sharded Data Parallel (FSDP)~\citep{fsdp} for efficient distributed training. During the exploration phase, response generation is accelerated using vLLM~\citep{vllm} via flash-attention~\citep{flashattention}. We utilize GRPO~\citep{grpo} as the RL algorithm. We set \emph{enable\_thinking} as \emph{False} for Qwen3 series to avoid the long thinking paradigm.

\paragraph{Domain-Specific Training Processes.}
Our experimental design validates our method across two distinct domains---Mathematics and Logic---spanning three model variants to assess generality across both model scale and model family: \textbf{Qwen3-1.7B}, \textbf{Qwen3-4B}~\citep{qwen3}, and \textbf{Llama-3.2-3B}~\citep{llama3}.

\textbullet\enspace For \textbf{Mathematics}, we perform RL directly on the \textbf{instruction-tuned} variants of all three base models. The reward mechanism is binary: a response is awarded a reward of 1.0 if the answer extracted from the \texttt{\textbackslash boxed\{\}} environment matches the ground truth, and 0.0 otherwise. This verification is performed using the \texttt{math-verify} library. We employ a relatively large number of rollouts per prompt ($n=16$) to provide stable variance estimates for policy gradient updates.

\textbullet\enspace For \textbf{Logic}, we likewise perform RL directly on the \textbf{instruction-tuned} variants of all three models, without any preceding supervised fine-tuning stage. The reward signal is binary and based on exact match: a response receives a reward of 1.0 if the prediction extracted from the \texttt{<answer>} tags exactly matches the ground-truth output, and 0.0 otherwise. We adopt the same number of rollouts per prompt ($n=16$) as in the Mathematics setting, maintaining consistency across domains.

\paragraph{Prompt.}
We employ domain-specific prompts tailored to the structure of each task type. In Mathematics, a consistent prompt format is applied to both training and evaluation. For Logic tasks, distinct prompt templates are used for each sub-task category (ARC-style grid tasks, ordering puzzles, and graph puzzles) to match the corresponding input format, with the same templates applied during both training and evaluation. The exact templates are shown below.

\begin{promptbox}[title=Mathematics Training \& Evaluation Prompt, label=box:math_prompt]
\footnotesize\ttfamily
\texttt{\textbf{System\_Prompt:} Please reason step by step, and put your final answer within \textbackslash\textbackslash boxed\{\}.}
\texttt{\textbf{User\_Prompt:} \{task\_text\}}
\end{promptbox}

\begin{promptbox}[title=Logic-ARC Training \& Evaluation Prompt, label=box:logic_arc_prompt]
\footnotesize\ttfamily
\texttt{\textbf{System\_Prompt:} You are a world-class puzzle solver with exceptional pattern recognition skills. Your task is to analyze puzzles, spot patterns, and provide direct solutions. Given input-output grid pairs as reference examples, carefully observe the patterns to predict the output grid for new test input. Each pair follows the same transformation rule. Grids are 2D arrays. Here are the input and output grids for the reference examples: \{User\_Prompt\}. What is the output grid? Please put your answer within <answer> and </answer> tags, your final answer should be only the output grid (2d array).}

\texttt{\textbf{User\_Prompt:} \{task\_text\}}
\end{promptbox}

\begin{promptbox}[title=Logic-OrderingPuzzle Training \& Evaluation Prompt, label=box:logic_order_prompt]
\footnotesize\ttfamily
\texttt{\textbf{System\_Prompt:} Solve the following puzzle to determine the order of the \{User\_Prompt\} from left to right. \{User\_Prompt\}. Please put your answer within <answer> and </answer> tags, for example <answer> ['pigeon', 'sparrow', 'quail'] </answer>.}

\texttt{\textbf{User\_Prompt:} \{category\} and \{task\_text\}}
\end{promptbox}

\begin{promptbox}[title=Logic-GraphPuzzle Training \& Evaluation Prompt, label=box:logic_graph_prompt]
\footnotesize\ttfamily
\texttt{\textbf{System\_Prompt:} Given the following list of predicates: \{User\_Prompt\}. Respond with only the trait of the next step. Please put your answer within <answer> and </answer> tags, for example <answer> fdebme </answer>.}

\texttt{\textbf{User\_Prompt:} \{task\_text\}}
\end{promptbox}

\paragraph{Hyperparameters.}
All three model variants share identical training hyperparameters within each domain; differences arise only between domains due to their distinct input length distributions and dataset sizes. Specifically, Logic tasks require substantially longer prompt contexts owing to the structured, multi-example format of ARC-style inputs, whereas Mathematics tasks demand longer response budgets to accommodate extended chain-of-thought reasoning. A comprehensive summary is provided in Table~\ref{tab:hyperparams}.

\begin{table*}[h!]
\centering
\resizebox{0.95\textwidth}{!}{%
\begin{tabular}{@{}lcc@{}}
\toprule
& \textbf{Mathematics (RL)} & \textbf{Logic (RL)} \\
\midrule
Base Model & Qwen3-\{1.7B, 4B\}. / Llama-3.2-3B. & Qwen3-\{1.7B, 4B\}. / Llama-3.2-3B. \\
\midrule
\multicolumn{3}{l}{\textbf{\textit{Common Training Parameters}}} \\
Optimizer & AdamW & AdamW \\
Learning Rate & $1 \times 10^{-6}$ & $1 \times 10^{-6}$ \\
Weight Decay & 0.1 & 0.1 \\
Gradient Clipping & 1.0 & 1.0 \\
Batch Size (Prompts) & 64 & 64 \\
Total Epochs & 1 & 3 \\
\midrule
\multicolumn{3}{l}{\textbf{\textit{RL-Specific Parameters}}} \\
Rollouts per Prompt ($n$) & 16 & 16 \\
GRPO Clip Ratio ($\epsilon$) & 0.2 & 0.2 \\
Rollout Temperature & 1.0 & 1.0 \\
\midrule
\multicolumn{3}{l}{\textbf{\textit{System and Memory Parameters}}} \\
Max Prompt Length & 4{,}096 & 28{,}672 \\
Max Response Length & 8{,}192 & 4{,}096 \\
Rollout Engine Number & 4 & 4 \\
Evaluation Steps Interval & 12 & 8 \\
\bottomrule
\end{tabular}%
}
\caption{Key hyperparameters for RL training across domains. All three model variants (Qwen3-1.7B, Qwen3-4B, Llama-3.2-3B) share the same configuration within each domain.}
\label{tab:hyperparams}
\end{table*}

\subsection{Policy Optimization Algorithm: GRPO}
\label{appendix:algorithm_grpo}

We adopt \textbf{Group Relative Policy Optimization (GRPO)}~\citep{grpo} as our policy gradient algorithm. A defining characteristic of GRPO is its elimination of a learned value function: rather than relying on a separate critic network to estimate baselines, it constructs the advantage signal directly from intra-group reward comparisons. Concretely, the policy parameters $\theta$ are optimized via a clipped surrogate objective of the form:
\begin{align}
\mathcal{L}_{\text{GRPO}}(\theta)
= {} & \mathbb{E}_{y \sim \pi_{\theta_{\text{old}}}} \Bigl[
\min \bigl( r_\theta(y) A(y), \nonumber\\
& \qquad \operatorname{clip}(r_\theta(y), 1-\epsilon, 1+\epsilon)\, A(y) \bigr)
\Bigr].
\label{eq:grpo_objective}
\end{align}

where $r_\theta(y) = {\pi_\theta(y|x)}/{\pi_{\theta_{\text{old}}}(y|x)}$ is the probability ratio measuring how the likelihood of response $y$ shifts between the updated and reference policies, and the clipping operator constrains this ratio to the interval $[1-\epsilon,\, 1+\epsilon]$, bounding each gradient step within a trusted region to prevent destabilizing policy updates.

The advantage $A(y_k)$ in Equation~\ref{eq:grpo_objective} is derived through a group normalization procedure. For each prompt $x$, a group of $G$ responses $Y = \{y_1, \dots, y_G\}$ is sampled from the old policy $\pi_{\theta_{\text{old}}}$, yielding a corresponding set of rewards $\{R(y_k)\}_{k=1}^G$. The advantage of each response is then defined as its reward standardized by the group statistics:
\begin{equation}
\label{eq:grpo_advantage}
A(y_k) = \frac{R(y_k) - \mu_Y}{\sigma_Y},
\end{equation}
where $\mu_Y$ and $\sigma_Y$ denote the mean and standard deviation of rewards within the group, respectively. This normalization ensures that the advantage signal reflects a response's relative quality among its peers, providing a low-variance baseline without requiring any additional learned components.

\subsection{Baselines}
\label{appendix:baselines}

To evaluate \textsc{TrajVal}, we select three representative online task-valuation methods as baselines. For each, we report results both with and without \textsc{TrajVal} applied as a sampling-weight multiplier. All baselines share the same base training configuration described in Appendix~\ref{appendix:train_details}; only the task-selection logic differs. Baseline-specific hyperparameters are set according to the recommendations of the respective original works.

\begin{itemize}[leftmargin=*]

    \item \textbf{\textsc{Base-GRPO}}~\citep{grpo}: The vanilla GRPO algorithm with uniform random task sampling, serving as the primary reference point. Tasks are drawn \textbf{randomly} from the training pool without any curriculum or difficulty-aware selection. This baseline establishes the performance floor that all task-selection methods, including \textsc{TrajVal}, are expected to improve upon.

    \item \textbf{\textsc{BOTS}}~\citep{bots}: \textbf{B}ayesian \textbf{O}nline \textbf{T}ask \textbf{S}election is a unified framework that recasts dynamic task selection as a Bayesian inference problem over evolving model capabilities. In our experiments, BOTS is initialized with pre-computed pass rates from two reference models (\texttt{Qwen2.5-7B} and \texttt{Qwen3-32B}) as the initial difficulty features. We follow the recommended configuration from the original work, with the key task-selector hyperparameters set as: group size $m=16$, implicit evidence weight $\lambda=0.1$, regularization coefficient $\rho=0.1$, target difficulty $p^{*}=0.5$, and Thompson sampling temperature $\tau=0.5$.

    \item \textbf{\textsc{GRESO}}~\citep{greso}: \textbf{G}RPO with \textbf{E}fficient \textbf{S}elective \textbf{R}ollout is an online pre-rollout filtering algorithm motivated by the empirical observation of strong temporal consistency in prompt informativeness: prompts that yield near-zero advantage in one training epoch are likely to remain uninformative in subsequent epochs. We adopt the recommended configuration from the original paper, with the key task-selector hyperparameters set as: easy-skipping probability $p_{\text{easy}}=0.5$, hard-skipping probability $p_{\text{hard}}=0.5$, zero-variance target $\sigma^2_{\text{zero}}=0.25$, probability update step $\Delta p=0.01$, and probability bounds $p_{\min}=0.05$, $p_{\max}=0.95$.

\end{itemize}

%% file: section/Appendix/extended_exps.tex
\section{Extended Experiments and Analyses}

\subsection{Design Intuition of Learnability Score}
\label{app:score_intuition}

Equation~\ref{eq:trajval-score} is intended as a low-cost operationalization of the two empirical dimensions introduced in Section~\ref{sec:heterogeneity}: \emph{capacity to improve} and \emph{constructive directional response}. The factor $1-\bar{r}_{\mathrm{early}}(x)$ down-weights tasks that are already near saturation, while $\max(0,\Delta(x))$ suppresses tasks whose reward trajectories are flat or regressive under optimization.

The multiplicative form reflects a conservative conjunctive design: a task should receive a high score only when it satisfies both conditions simultaneously. If either dimension is weak---for example, a task is already nearly mastered, or its trajectory shows little positive response---the overall score is reduced accordingly. This behavior matches the qualitative motivation in Figure~\ref{fig:traj_main}, where tasks with meaningful training value tend to exhibit both remaining headroom and a non-negative optimization trend.

We do not claim that this product form is unique or theoretically optimal. Rather, it provides a simple and stable proxy for prioritizing tasks that are both unsaturated and empirically responsive to training, while avoiding additional estimation complexity. The ablations in Section~\ref{sec:ablation} further support this choice by showing that combining the two factors is more effective than using either one alone.

\subsection{Stability of Learnability Profiles}
\label{app:stability_profiles}

This subsection provides the full visualization for the cross-context stability study summarized in Section~\ref{sec:empirical_case}. We train Qwen3-1.7B for 20 epochs on the full pool $\mathcal{D}$ and on two independently sampled subsets $\mathcal{P}_1$ and $\mathcal{P}_2$, each of size 512, with 256 shared tasks. For each training context, we assign every task a trajectory profile using the criteria in Figure~\ref{fig:traj_main}.

\begin{figure}[h]
    \centering
    \includegraphics[width=\linewidth]{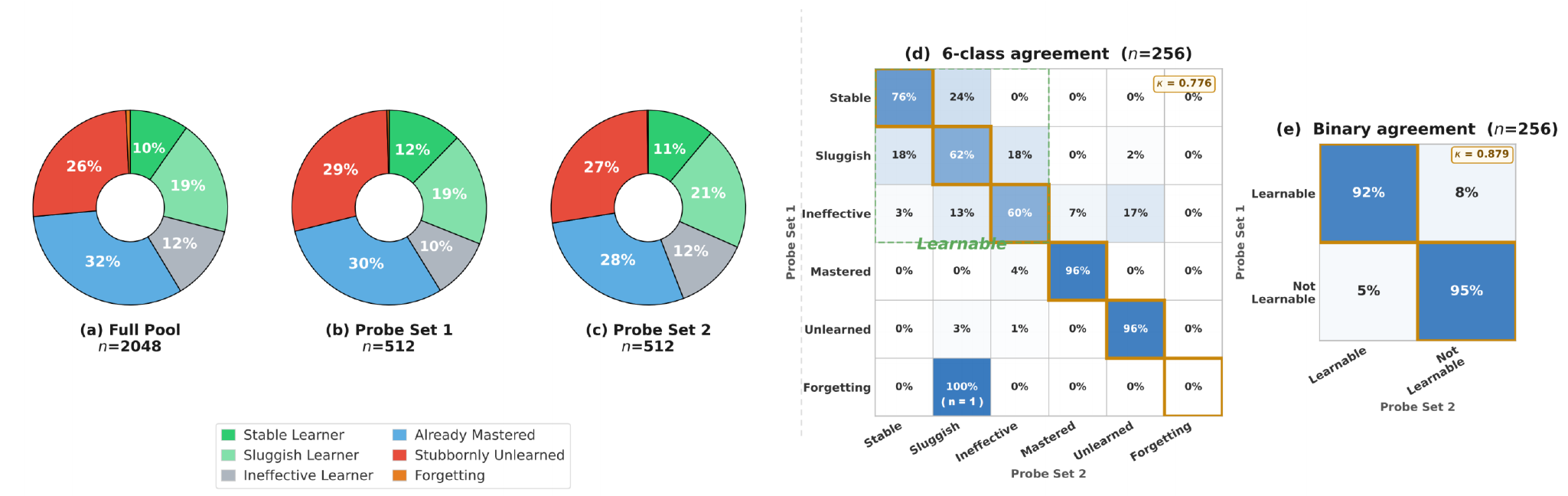}
    \caption{Qwen3-1.7B: Stability of learnability profiles across independently sampled training contexts. \textbf{(a)--(c):} The macroscopic distribution of profiles remains nearly identical across $\mathcal{D}$, $\mathcal{P}_1$, and $\mathcal{P}_2$, confirming that uniform subsampling preserves the learnability composition of the full pool. \textbf{(d)--(e):} Confusion matrices for the 256 shared tasks, evaluated under $\mathcal{P}_1$ versus $\mathcal{P}_2$. Strong per-task agreement is observed at both the fine-grained six-profile level ($\kappa = 0.776$) and the binary learnable/non-learnable level ($\kappa = 0.879$).}
    \label{fig:stability_qwen3}
\end{figure}

Figure~\ref{fig:stability_qwen3}(a)--(c) examines stability at the pool level. Despite the substantial difference in set size and the independent sampling of $\mathcal{P}_1$ and $\mathcal{P}_2$, the profile proportions remain highly similar across all three contexts. In particular, the aggregate mass of learnable profiles and non-learnable profiles changes only slightly, and the relative ordering of the major categories is preserved. This supports the view that learnability composition is largely a property of the underlying data source, rather than a fragile consequence of a specific training mixture.

Figure~\ref{fig:stability_qwen3}(d)--(e) examines stability at the level of individual tasks. Among the 256 shared tasks, most assignments lie on or near the diagonal of the confusion matrices. Agreement is already strong for the full six-class taxonomy, with most disagreement occurring between neighboring learnable categories such as \textsc{Stable}, \textsc{Sluggish}, and \textsc{Ineffective}. After collapsing profiles into the binary learnable/non-learnable partition, the consistency becomes even stronger, yielding $\kappa=0.879$. This pattern is expected: fine-grained labels are more sensitive to small variations in trajectory shape, whereas the coarser partition captures the directional distinction that matters most for data allocation.

Taken together, the figure supports two practical conclusions used by \textsc{TrajVal}. First, a uniformly sampled subset can preserve the broad learnability composition of the full pool. Second, per-task learnability is sufficiently stable across contexts to serve as a reusable signal for probe-based estimation.

\subsection{Learnability Stability Across Model Families on Llama}
\label{app:stability_family}

To assess whether the observations in Sections~\ref{sec:learnability}--\ref{sec:main} extend beyond the Qwen family, we repeat the key analyses on Llama-3.2-3B. The goal of this subsection is not to introduce a new setting, but to verify that the main empirical patterns remain visible under a different base model family.

\begin{figure}[h]
    \centering
    \includegraphics[width=\linewidth]{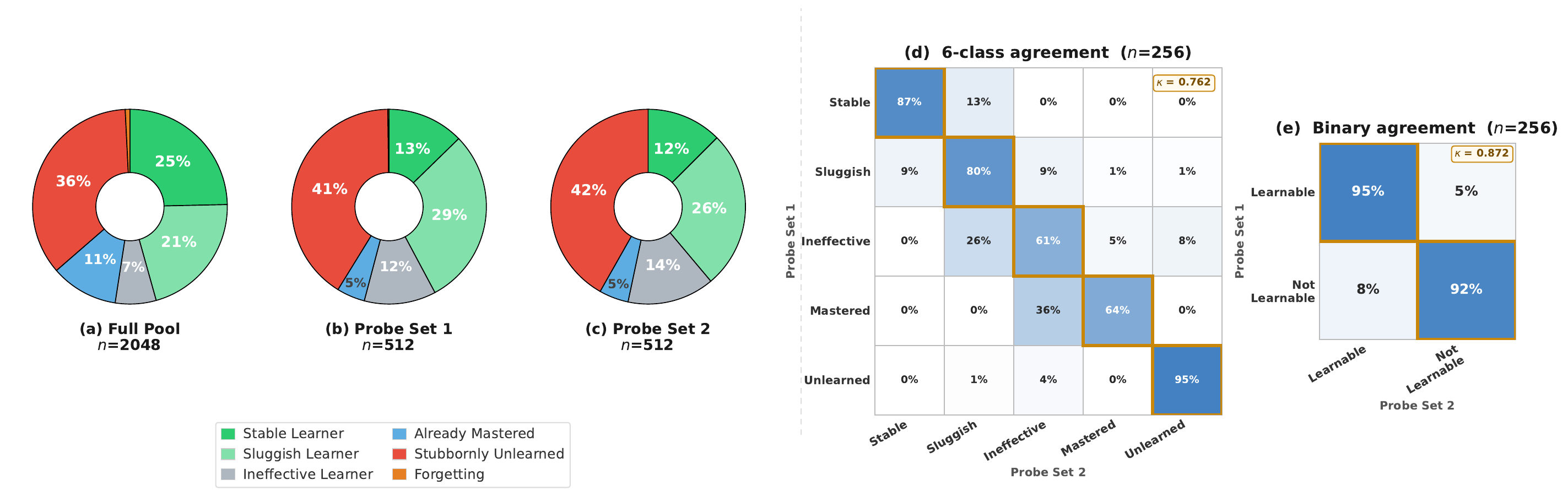}
    \caption{Llama-3.2-3B: Stability of learnability profiles across independently sampled training contexts. The configuration is the same as in Figure \ref{fig:stability_qwen3}.}
    \label{fig:stability_llama}
\end{figure}

\paragraph{Cross-context stability.}
We first examine whether learnability remains stable across training contexts, as in the Qwen-based study of Section~\ref{sec:empirical_case}. Figure~\ref{fig:stability_llama} shows that the macroscopic profile distribution remains similar across independently sampled training contexts, and that the profile assignment of shared tasks is stable at both the six-profile level and the binary learnable/non-learnable level. The resulting agreement scores, $\kappa = 0.762$ for the six-way taxonomy and $\kappa = 0.872$ for the binary grouping, are close to those observed in the main text. This suggests that learnability remains a measurable and reasonably stable task property under a different model family.

\begin{figure}[h]
    \centering
    \includegraphics[width=0.95\linewidth]{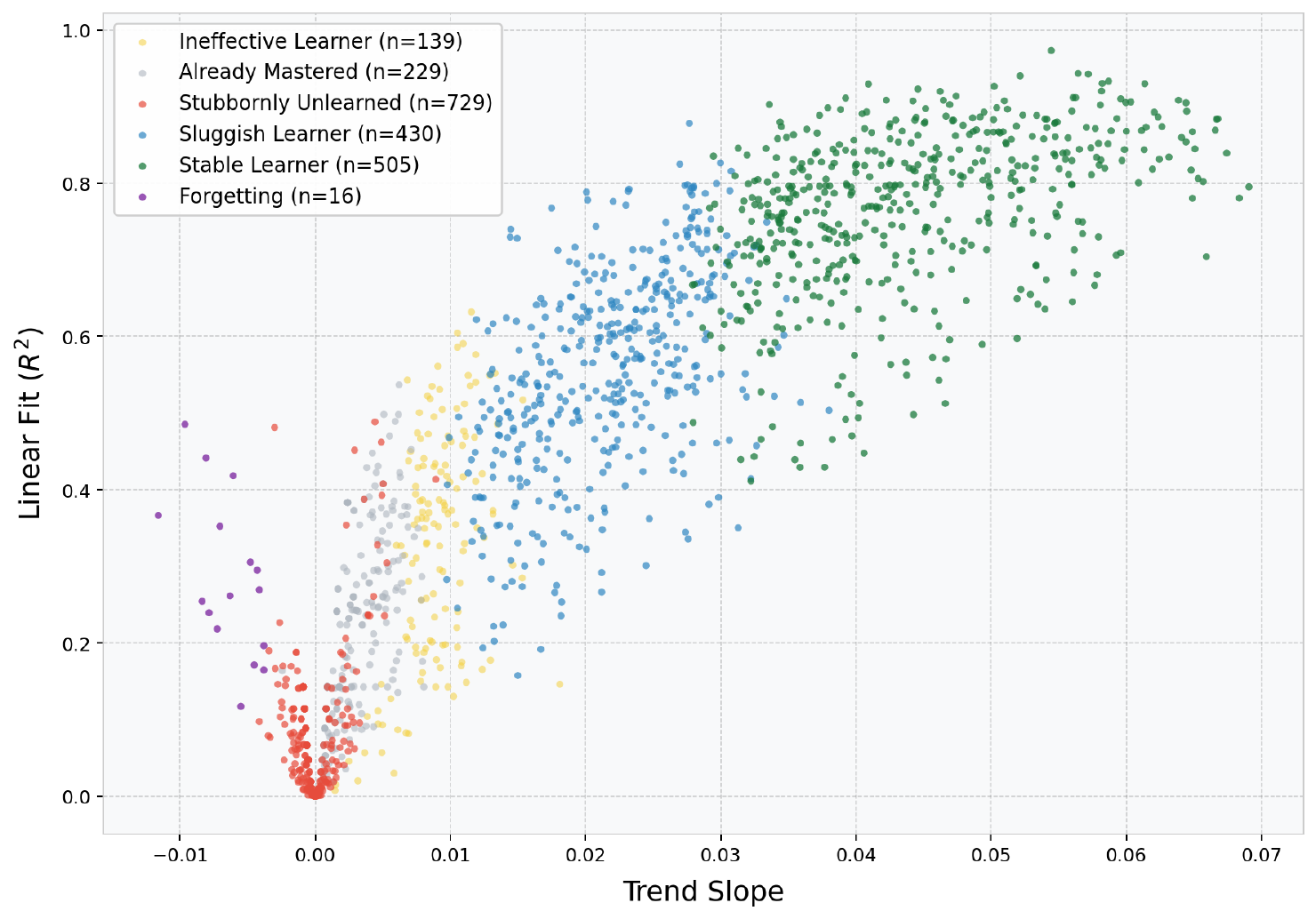}
    \caption{Llama-3.2-3B: All tasks in $\mathcal{D}$ projected onto the slope--$R^2$ space of their reward trajectories, colored by profile labels derived from the endpoint pair $(\bar{r}_{\mathrm{early}}, \bar{r}_{\mathrm{late}})$. The same broad structural separation observed in Figure~\ref{fig:slope_r2} is preserved, indicating that endpoint-derived labels continue to recover the global organization of trajectory behavior under a different model family.}
    \label{fig:app_slope_r2}
\end{figure}

\paragraph{Learnability structure.}
We next revisit whether the structural organization of learnability observed in Section~\ref{sec:learnability} also appears under Llama-3.2-3B. Figure~\ref{fig:app_slope_r2} reproduces the slope--$R^2$ projection on the full 2,048-task pool, with profile labels derived from the endpoint pair $(\bar{r}_{\mathrm{early}}, \bar{r}_{\mathrm{late}})$. The same broad geometric structure is preserved: positively trending profiles occupy the high-slope region, while non-learnable profiles remain concentrated near low or negative slopes. This indicates that the endpoint summary continues to recover the global organization of trajectory behavior under Llama-3.2-3B, consistent with the reduction in Section~\ref{sec:method}.

\paragraph{End-to-end transfer.}
Finally, we ask whether the practical benefits of the learnability prior also transfer to downstream RL training. Table~\ref{tab:llama} reports the end-to-end results on mathematical and logical reasoning. The overall pattern is directionally consistent with the main text: \textsc{TrajVal} usually improves either convergence speed, final accuracy, or AUC when paired with Random or BOTS, and remains particularly helpful on Logic. At the same time, the gains are smaller and less uniform than those observed for Qwen, with one Math configuration showing a regression when combined with GRESO. We therefore view the Llama results as supportive rather than definitive: they suggest that the learnability prior transfers across model families, while also indicating that its interaction with specific schedulers can remain model-dependent.

\input{section/Tables/appendix_llama_results}

\subsection{Diagnostic Analysis of TrajVal Scores}
\label{appendix:diagnostic}

\paragraph{Score distribution across learnability profiles.}
To verify that the continuous score $s(x)$ assigns weight in a manner consistent with the qualitative learnability taxonomy introduced in Section~\ref{sec:learnability}, we stratify $\mathcal{D}$ by trajectory profile and report per-profile score statistics under two independently sampled probe sets $\mathcal{P}_1$ and $\mathcal{P}_2$ in Table~\ref{tab:score_distribution}. Three patterns are evident. First, the learnable profiles collectively receive $67\%$ of total sampling weight despite constituting only $41\%$ of the pool, representing a $1.6\times$ redirection of sampling budget toward tasks with positive learnability. Second, score statistics within each group respect the expected learnability ordering: mean $s(x)$ decreases monotonically from \textsc{Stable Learner} ($0.521$) through \textsc{Sluggish Learner} ($0.470$) and \textsc{Ineffective Learner} ($0.391$), while the fraction of tasks collapsed to the probability floor rises correspondingly, reaching $92\%$ for \textsc{Stubbornly Unlearned}. Third, distributions under $\mathcal{P}_1$ and $\mathcal{P}_2$ are nearly identical across all profiles, confirming that the score reflects a stable property of the data rather than an artifact of any particular probe realization.

\begin{table*}[h]
\centering
\setlength{\tabcolsep}{5pt}
\begin{tabular}{lccccccc}
\toprule
& & \multicolumn{3}{c}{$\mathcal{P}_1$} & \multicolumn{3}{c}{$\mathcal{P}_2$} \\
\cmidrule(lr){3-5}\cmidrule(lr){6-8}
\textbf{Profile} & $n$ & Share & Mean & Floor\% & Share & Mean & Floor\% \\
\midrule
\textsc{Stable Learner}           & 200  & 18.0\% & 0.521 & 30.5\% & 16.3\% & 0.456 & 39.0\% \\
\textsc{Sluggish Learner}         & 394  & 32.0\% & 0.470 & 30.5\% & 31.9\% & 0.454 & 32.0\% \\
\textsc{Ineffective Learner}      & 253  & 17.1\% & 0.391 & 31.2\% & 17.4\% & 0.386 & 31.6\% \\
\textsc{Already Mastered}         & 659  & 20.7\% & 0.182 & 63.3\% & 21.4\% & 0.182 & 63.6\% \\
\textsc{Stubbornly Unlearned}     & 527  & 11.7\% & 0.129 & 92.4\% & 12.1\% & 0.129 & 93.6\% \\
\textsc{Forgetting}               &  15  &  0.5\% & 0.210 & 73.3\% &  0.9\% & 0.339 & 53.3\% \\
\midrule
\textsc{Learnable} (combined)     & 847  & 67.0\% & 0.459 & 30.7\% & 65.6\% & 0.434 & 33.5\% \\
\textsc{Non-learnable} (combined) & 1201 & 33.0\% & 0.159 & 76.2\% & 34.4\% & 0.160 & 76.6\% \\
\bottomrule
\end{tabular}
\caption{Per-profile distribution of \textsc{TrajVal} scores under two independent probe sets. \textbf{Share}: fraction of total sampling weight. \textbf{Mean}: mean $s(x)$ within profile. \textbf{Floor\%}: fraction of tasks at the probability floor.}
\label{tab:score_distribution}
\end{table*}

\paragraph{Score stability under varying probe horizons.}
\textsc{TrajVal} derives its score from $\theta_\mathrm{probe}$ obtained after $T$ epochs of probe training. We examine how early the resulting score ordering stabilizes by saving intermediate checkpoints $\theta_\mathrm{probe}^{(k)}$ at each epoch $k \in \{1, \ldots, T\}$ ($T=20$), recomputing $\mathrm{post}_k(x)$ over the full pool $\mathcal{D}$, and measuring the Spearman rank correlation $\rho$ between the intermediate score $s_k(x) = (1-\mathrm{pre}(x)) \cdot \max(0, \mathrm{post}_k(x) - \mathrm{pre}(x))$ and the final score $s_T(x)$.

\begin{figure}[h]
    \centering
    \includegraphics[width=0.95\linewidth]{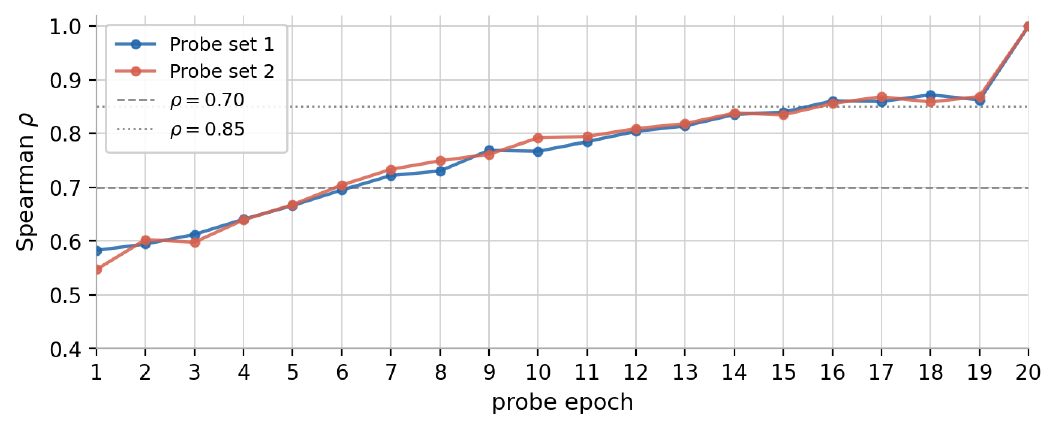}
    \caption{Spearman rank correlation between $s_k(x)$ and $s_T(x)$ as a function of probe epoch $k$, evaluated on $\mathcal{D}$. Two independently sampled probe sets are shown. The ordering stabilizes ($\rho \geq 0.85$) by $k=14$, well before the full probe horizon $T=20$.}
    \label{fig:probe_epoch_stability}
\end{figure}

As shown in Figure~\ref{fig:probe_epoch_stability}, $\rho$ rises rapidly in the early epochs and reaches $0.85$ by $k=14$, well before the nominal probe horizon $T=20$. Beyond $k=14$, the correlation remains stable and the two independently drawn probe sets stay within $0.04$ of each other at all epochs. This indicates that the relative learnability ordering captured by \textsc{TrajVal} crystallizes early in probe training, and that the probe budget could be reduced to roughly $70\%$ of the nominal horizon without meaningfully degrading score quality.

\subsection{Computational Cost of TrajVal}
\label{appendix:compute_cost}

\paragraph{Cost Structure.}
The preprocessing overhead of \textsc{TrajVal} decomposes into two components. The dominant cost is the probe training run, in which the base model $\theta_0$ is fine-tuned via RL on a uniformly sampled probe set $\mathcal{P} \subset \mathcal{D}$. In our primary configuration, $|\mathcal{P}| = 512$ for a pool of $|\mathcal{D}| = 17{,}398$ tasks, corresponding to approximately $3\%$ of the full training pool; the total probe training cost is therefore $O(T \cdot |\mathcal{P}|)$, which is a small fraction of the main training budget by construction. The secondary cost is endpoint estimation: for every task $x \in \mathcal{D}$, we evaluate $\bar{r}_{\mathrm{early}}(x)$ and $\bar{r}_{\mathrm{late}}(x)$ via two targeted inference passes over the full pool using the recorded probe checkpoints, incurring $O(2 \cdot |\mathcal{D}|)$ forward passes with no gradient computation. Crucially, once the per-task sampling weights $\{p(x)\}$ are computed, \textsc{TrajVal} introduces \emph{no additional overhead during the main training phase}: the weights are fixed prior to training, and sampling at each step reduces to a single weighted multinomial draw.

\paragraph{Empirical Overhead Measurement.}
Total training time is dominated by model rollout length---a factor orthogonal to our method---we focus on the \emph{experience pipeline processing time} (\texttt{time/experience\_pipeline/total}) as the primary measurement target. This metric captures the end-to-end latency of the data-processing pipeline at each training step, encompassing experience deserialization, advantage computation, and output buffer writes, and is the only pipeline-level signal that can reflect task-selector overhead in our logging infrastructure. It explicitly excludes model inference and parameter update time, making it a clean proxy for the marginal cost.

\begin{table*}[h]
\centering
\vspace{0.5em}
\begin{subtable}{\linewidth}
\centering
\begin{tabular}{lcccc}
\toprule
\textbf{Statistic} & \textbf{Random} & \textbf{\textsc{TrajVal}} & \textbf{Abs.\ Diff.} & \textbf{Change} \\
\midrule
Mean & 2.301\,s & 2.247\,s & $-$0.055\,s & $-$2.4\% \\
Median & 2.286\,s & 2.304\,s & $+$0.018\,s & $+$0.8\% \\
p90 & 2.957\,s & 2.971\,s & $+$0.014\,s & $+$0.5\% \\
p95 & 3.177\,s & 3.173\,s & $-$0.003\,s & $-$0.1\% \\
Cumulative (271 steps) & 623.6\,s & 608.9\,s & $-$14.8\,s & $-$2.4\% \\
\bottomrule
\end{tabular}
\caption{Experience pipeline total time (\texttt{time/experience\_pipeline/total}).}
\end{subtable}

\vspace{0.8em}

\begin{subtable}{\linewidth}
\centering
\begin{tabular}{lcccc}
\toprule
\textbf{Statistic} & \textbf{Random} & \textbf{\textsc{TrajVal}} & \textbf{Abs.\ Diff.} & \textbf{Pipeline Share} \\
\midrule
Mean latency & 0.019\,ms & 0.239\,ms & $+$0.220\,ms & 0.010\% \\
p99 latency & 0.033\,ms & 0.253\,ms & $+$0.221\,ms & 0.011\% \\
Cumulative (271 steps) & 5.2\,ms & 64.8\,ms & $+$59.4\,ms & 0.010\% \\
\bottomrule
\end{tabular}
\caption{Micro-benchmark: \texttt{get\_indices()} latency over 1{,}000 independent calls; pipeline share computed relative to the 623.6\,s random baseline.}
\end{subtable}
\caption{Empirical overhead of \textsc{TrajVal} relative to random sampling. \textbf{Top}: Experience pipeline processing time over 271 training steps. Mean and percentile differences are within measurement noise. \textbf{Bottom}: Micro-benchmark isolating \texttt{operator} latency over 1{,}000 calls. Although the relative increase appears large, the absolute per-step cost is equivalent to $0.01\%$ of total pipeline time.}
\label{tab:compute_cost}
\end{table*}

Table~\ref{tab:compute_cost} (top) reports the pipeline-level measurements over 271 training steps. Across the reported percentiles up to p95, the difference between \textsc{TrajVal} and random sampling remains below 0.02\,s, well within the variability attributable to system-level noise such as network I/O and serialization. The cumulative pipeline times are statistically indistinguishable. To further isolate the selector's contribution, Table~\ref{tab:compute_cost} (bottom) presents a direct micro-benchmark of the \texttt{Trinity-Operator} call. \textsc{TrajVal}'s weighted multinomial sampling takes approximately 0.24\,ms per call, compared to 0.02\,ms for uniform random sampling. While the relative difference is large in percentage terms, the absolute per-step overhead is 0.22\,ms---accumulating to roughly 59\,ms over 271 steps, or $0.01\%$ of the total pipeline time. This confirms that the learnability-aware sampling logic is entirely subsumed by other pipeline components and does not constitute a measurable bottleneck in practice.

\paragraph{Cost-Benefit Summary.}
The marginal cost of \textsc{TrajVal}---a one-time probe run on $\approx 3\%$ of the training pool, two inference passes for endpoint estimation, and a negligible per-step selector overhead---is offset by substantial gains in training efficiency. As reported in Section~\ref{sec:main}, \textsc{TrajVal}-augmented methods consistently reduce the Steps-to-Baseline (S2B) metric across all evaluated configurations, with S2B as low as $40.0\%$ on Logic at the 4B scale. This indicates that the same final performance level is achievable in less than half the training steps, implying a net computational gain that far exceeds the preprocessing overhead introduced by \textsc{TrajVal}.

\subsection{Cross-Regime Reuse of the Learnability Prior}
\label{sec:transfer}

Since learnability is regime-conditional, estimating the prior on the target
regime itself (\emph{in-regime}) is our default and most accurate setting. As
re-running the probe for every regime repeats the one-time preprocessing cost of
Appendix~\ref{appendix:compute_cost}, we ask whether an existing prior can be
reused: (i)~\emph{same-family, cross-scale}, from a smaller to a larger model in
the same family; and (ii)~\emph{cross-family}, from a different family. All runs
are on Math; each transferred prior guides RL on the target without
re-estimation, against uniform sampling (Random / \textsc{Base-GRPO}) on that
target, with the in-regime result on Qwen3-4B as a reference.

\begin{table*}[t]
\centering
\small
\begin{tabular}{llccc}
\toprule
Target & Prior source & Best Acc$\uparrow$ & AUC$\uparrow$ & S2B$\downarrow$ \\
\midrule
\multicolumn{5}{l}{\textit{Qwen3-4B}} \\
\quad Random          & --                        & 0.5060 & 0.4829 & --   \\
\quad + \textsc{TrajVal} (transfer)   & Qwen3-1.7B    & 0.5180 & 0.4889 & 66.0 \\
\quad + \textsc{TrajVal} (transfer)   & Llama-3.2-3B  & 0.5129 & 0.4817 & 84.0 \\
\quad + \textsc{TrajVal} (\emph{in-regime}) & Qwen3-4B & \textit{0.5190} & \textit{0.4931} & \textit{60.0} \\
\midrule
\multicolumn{5}{l}{\textit{Qwen3-8B}} \\
\quad Random          & --                        & 0.5342 & 0.4999 & --   \\
\quad + \textsc{TrajVal} (transfer)   & Qwen3-4B      & 0.5484 & 0.5109 & 87.5 \\
\bottomrule
\end{tabular}
\caption{Reusing the \textsc{TrajVal} prior across regimes (Math). Each
transferred prior is estimated on the \emph{prior source} and guides RL on the
\emph{target} without re-estimation. S2B is the fraction of steps to reach the
Random baseline's peak accuracy (lower is better; ``--'' marks that baseline).
The in-regime row is a reference, not a competing method.}
\label{tab:transfer}
\end{table*}

\paragraph{Same-family, cross-scale reuse.}
Within Qwen3, a prior from a smaller model remains useful on a larger one, improving
all three metrics over uniform sampling: Qwen3-1.7B$\to$Qwen3-4B reaches the baseline
peak in 66.0\% of steps, and Qwen3-4B$\to$Qwen3-8B in 87.5\% (Table~\ref{tab:transfer}).
A single prior can thus be shared across scales rather than recomputed per model.

\paragraph{Cross-family reuse.}
Transfer across families also helps, but less. Llama-3.2-3B$\to$Qwen3-4B raises Best
Acc (0.5060$\to$0.5129) and reaches the peak in 84.0\% of steps, while AUC is
essentially unchanged (0.4817 vs.\ 0.4829). Both the smaller Best Acc gain and the
flat AUC are consistent with a prior less matched to the target regime.

\paragraph{}
Overall, the in-regime prior on Qwen3-4B is strongest (Best Acc 0.5190), slightly
above same-family (0.5180) and cross-family (0.5129) transfer; same-family reuse
stays closest to it and converges fastest (S2B 66.0 vs.\ 84.0). In-regime estimation
therefore remains the default for peak performance, while transfer is a lower-cost
alternative that spreads the probe cost across models---same-family reuse being the
more dependable of the two.

\input{section/Tables/appendix_math_complete_results.tex}

\subsection{Hyper-parameter Sensitivity: Power Compression Exponent}
\label{appendix:hyperparam}

Table~\ref{tab:ablation_alpha} reports the sensitivity of \textsc{TrajVal} to the power
compression exponent $\alpha$, which controls the degree to which learnability scores
are flattened before conversion to sampling probabilities. Across three values
$\alpha \in \{0.1, 0.3, 0.5\}$, the overall pattern of improvement over the
\textsc{Random} baseline is preserved on both domains. On Math, all three settings
advance Best Acc beyond the random baseline ($0.3967$), and $\alpha = 0.3$
achieves the most favorable convergence speed, reaching the random baseline's peak
accuracy in only $60.6\%$ of the steps while maintaining a competitive AUC of
$0.3729$. On Logic, the Best Acc increases monotonically with $\alpha$ from $0.1884$
to $0.1924$, while $\alpha = 0.3$ yields the highest AUC ($0.1663$) among all
variants, suggesting that a moderate compression level best balances priority
concentration and distributional coverage.

Taken together, these results indicate that \textsc{TrajVal} is not sensitive to the
precise choice of $\alpha$ within the range evaluated: all three settings produce
non-trivial improvements over uniform sampling on both domains and both metrics,
and no single value of $\alpha$ degrades performance below the \textsc{Random} baseline.
The moderate value $\alpha = 0.3$ provides the best overall trade-off and is therefore
adopted as the default throughout the main experiments.

\begin{table*}[h!]
\centering
\setlength{\tabcolsep}{5pt}
\begin{tabular}{lcccccc}
\toprule
\small
& \multicolumn{3}{c}{Math (Avg)}
& \multicolumn{3}{c}{Logic} \\
\cmidrule(lr){2-4}\cmidrule(lr){5-7}
\textbf{Method} & S2B (\%) $\downarrow$ & Best Acc $\uparrow$ & AUC $\uparrow$
                & S2B (\%) $\downarrow$ & Best Acc $\uparrow$ & AUC $\uparrow$ \\
\midrule
Random                      & --    & 0.3967 & 0.3620 & --    & 0.1826 & 0.1544 \\
\midrule
TrajVal ($\alpha = 0.1$)    & 78.79 & 0.4062 & 0.3656 & 69.57 & 0.1884 & 0.1611 \\
TrajVal ($\alpha = 0.3$)    & 60.61 & 0.4035 & 0.3729 & 73.91 & 0.1904 & 0.1663 \\
TrajVal ($\alpha = 0.5$)    & 78.79 & 0.4071 & 0.3647 & 69.57 & 0.1924 & 0.1660 \\
\bottomrule
\end{tabular}
\caption{Sensitivity to the power compression exponent $\alpha$ on Qwen3-1.7B. All other hyperparameters are held fixed.}
\label{tab:ablation_alpha}
\end{table*}

%% file: section/Tables/appendix_llama_results.tex
\begin{table*}[h]
\centering
\setlength{\tabcolsep}{5pt}
\begin{tabular}{lcccccc}
\toprule
& \multicolumn{3}{c}{Math (Avg)}
& \multicolumn{3}{c}{Logic} \\
\cmidrule(lr){2-4}\cmidrule(lr){5-7}
\textbf{Method} & S2B (\%) $\downarrow$ & Best Acc $\uparrow$ & AUC $\uparrow$
                & S2B (\%) $\downarrow$ & Best Acc $\uparrow$ & AUC $\uparrow$ \\
\midrule
Random         & --     & 0.1724 & 0.1582 & --     & 0.1204 & 0.1031 \\
Random+TrajVal & \textbf{94.12}  & \textbf{0.1800} & \textbf{0.1637} & 100.00 & \textbf{0.1234} & \textbf{0.1059} \\
\midrule
BOTS           & --     & 0.1752 & 0.1585 & --     & 0.1332 & 0.1055 \\
BOTS+TrajVal   & \textbf{100.00} & \textbf{0.1760} & \textbf{0.1595} & \textbf{96.15}  & \textbf{0.1340} & \textbf{0.1094} \\
\midrule
GRESO          & --     & 0.1765 & \textbf{0.1646} & --     & 0.1252 & 0.1013 \\
GRESO+TrajVal  & \textbf{90.00}     & \textbf{0.1791} & 0.1634 & \textbf{65.00}  & \textbf{0.1288} & \textbf{0.1102} \\
\bottomrule
\end{tabular}
\caption{End-to-end results on mathematical and logical reasoning with Llama-3.2-3B. S2B (\%) denotes the fraction of baseline training steps required to match the baseline's peak performance, where lower is better. Best Acc is the peak evaluation accuracy over the full training run, and AUC is the area under the evaluation curve.}
\label{tab:llama}
\end{table*}

% \begin{table}[t]
% \centering
% \caption{Results on mathematical and logical reasoning, Llama3.2-3B.
% S2B (\%) denotes the fraction of baseline training steps required to match baseline peak performance (lower is better).
% Best Acc is the peak accuracy over the full training run.
% AUC is the area under the evaluation curve.
% All numbers are means over [K] independent seeds.}
% \label{tab:llama}
% \small
% \setlength{\tabcolsep}{5pt}
% \begin{tabular}{lcccccc}
% \toprule
% & \multicolumn{3}{c}{Math (Avg)}
% & \multicolumn{3}{c}{Logic} \\
% \cmidrule(lr){2-4}\cmidrule(lr){5-7}
% \textbf{Method} & S2B (\%) $\downarrow$ & Best Acc $\uparrow$ & AUC $\uparrow$
%                 & S2B (\%) $\downarrow$ & Best Acc $\uparrow$ & AUC $\uparrow$ \\
% \midrule
% Random         & --     & 0.1724 & 0.1582 & --     & 0.1204 & 0.1031 \\
% Random+TrajVal & \textbf{94.12}  & \textbf{0.1800} & \textbf{0.1637} & 100.00 & \textbf{0.1234} & \textbf{0.1059} \\
% \midrule
% BOTS           & --     & 0.1752 & 0.1585 & --     & 0.1332 & 0.1055 \\
% BOTS+TrajVal   & \textbf{100.00} & \textbf{0.1760} & \textbf{0.1595} & \textbf{96.15}  & \textbf{0.1340} & \textbf{0.1094} \\
% \midrule
% GRESO          & --     & \textbf{0.1793} & \textbf{0.1612} & --     & 0.1252 & 0.1013 \\
% GRESO+TrajVal  & --     & 0.1715 & 0.1587 & \textbf{65.00}  & \textbf{0.1288} & \textbf{0.1102} \\
% \bottomrule
% \end{tabular}
% \end{table}

%% file: section/Tables/appendix_math_complete_results.tex
\begin{table*}[h]
\centering
\setlength{\tabcolsep}{3pt}
\resizebox{0.9\textwidth}{!}{%
\begin{tabular}{lccccccccccccccccccc}
\toprule
& \multicolumn{3}{c}{MATH500}
& \multicolumn{3}{c}{Minerva}
& \multicolumn{3}{c}{Olympiad}
& \multicolumn{3}{c}{AMC23}
& \multicolumn{3}{c}{AIME25}
& \multicolumn{3}{c}{AIME24} \\
\cmidrule(lr){2-4}\cmidrule(lr){5-7}\cmidrule(lr){8-10}
\cmidrule(lr){11-13}\cmidrule(lr){14-16}\cmidrule(lr){17-19}
\textbf{Method}
& Best & AUC & S2B
& Best & AUC & S2B
& Best & AUC & S2B
& Best & AUC & S2B
& Best & AUC & S2B
& Best & AUC & S2B \\
\midrule
Random         & 75.5 & 72.1 & --   & 19.6 & 18.0 & --   & 43.2 & 39.9 & --   & 60.4 & 53.1 & --    & 17.6 & 15.3 & --    & 21.9 & 18.7 & --    \\
+Ours       & \textbf{76.3} & \textbf{72.9} & \textbf{75.8} & 19.5 & \textbf{18.4} & --   & 44.0 & \textbf{40.5} & \textbf{84.4} & \textbf{62.6} & \textbf{55.1} & \textbf{90.9} & \textbf{20.6} & \textbf{16.5} & \textbf{45.5} & \textbf{24.0} & \textbf{20.3} & \textbf{50.5} \\
\midrule
BOTS           & 75.5 & 72.1 & --   & 19.7 & 18.1 & --   & 44.1 & 39.9 & --   & 61.1 & 54.1 & --    & 18.8 & 15.7 & --    & 23.5 & 18.9 & --    \\
+Ours       & \textbf{76.6} & \textbf{72.9} & \textbf{94.4} & \textbf{20.0} & 18.1 & \textbf{95.0} & \textbf{44.9} & \textbf{40.9} & \textbf{85.0} & \textbf{62.4} & \textbf{54.9} & \textbf{95.0} & 18.5 & \textbf{15.9} & --    & 23.8 & \textbf{19.5} & \textbf{75.0} \\
\midrule
GRESO          & 75.3 & 71.7 & --   & 20.1 & 18.4 & --   & 44.1 & 39.7 & --   & 60.9 & 53.4 & --    & 18.2 & 15.4 & --    & 23.1 & 18.8 & --    \\
+Ours       & \textbf{76.0} & \textbf{72.6} & \textbf{85.0} & 19.7 & 18.3 & --   & \textbf{44.9} & \textbf{40.2} & \textbf{90.0} & \textbf{61.8} & \textbf{54.3} & \textbf{100.0}& \textbf{19.0} & \textbf{15.9} & \textbf{84.2} & \textbf{24.3} & \textbf{19.6} & \textbf{85.0} \\
\bottomrule
\end{tabular}%
}
\caption{Per-benchmark results on mathematical reasoning, Qwen3-1.7B. Best Acc ($\uparrow$), AUC ($\uparrow$), and S2B (\%, $\downarrow$) for each benchmark and method. Best results per column in \textbf{bold}.}
\label{tab:math_1b_full}
\end{table*}

\begin{table*}[h]
\centering
\footnotesize
\setlength{\tabcolsep}{3pt}
\resizebox{0.9\textwidth}{!}{%
\begin{tabular}{lccccccccccccccccccc}
\toprule
& \multicolumn{3}{c}{MATH500}
& \multicolumn{3}{c}{Minerva}
& \multicolumn{3}{c}{Olympiad}
& \multicolumn{3}{c}{AMC23}
& \multicolumn{3}{c}{AIME25}
& \multicolumn{3}{c}{AIME24} \\
\cmidrule(lr){2-4}\cmidrule(lr){5-7}\cmidrule(lr){8-10}
\cmidrule(lr){11-13}\cmidrule(lr){14-16}\cmidrule(lr){17-19}
\textbf{Method}
& Best & AUC & S2B
& Best & AUC & S2B
& Best & AUC & S2B
& Best & AUC & S2B
& Best & AUC & S2B
& Best & AUC & S2B \\
\midrule
Random         & 82.6 & 80.9 & --    & 24.8 & 23.5 & --    & 52.2 & 50.0 & --    & 79.5 & 75.8 & --    & 28.4 & 27.0 & --    & 36.3 & 32.6 & --    \\
+Ours          & \textbf{83.8} & \textbf{81.9} & \textbf{60.0} & 24.6 & \textbf{23.5} & --    & \textbf{53.8} & \textbf{51.0} & \textbf{60.0} & \textbf{80.9} & \textbf{77.8} & \textbf{60.0} & \textbf{31.3} & \textbf{28.6} & \textbf{43.6} & \textbf{37.6} & \textbf{32.9} & \textbf{72.0} \\
\midrule
BOTS           & 82.9 & 80.9 & --    & 24.4 & 23.4 & --    & 52.0 & 49.3 & --    & 80.5 & 75.0 & --    & 29.5 & 26.2 & --    & 35.9 & 31.2 & --    \\
+Ours          & \textbf{83.8} & \textbf{81.5} & \textbf{70.0} & \textbf{24.8} & \textbf{23.5} & \textbf{94.4} & \textbf{53.5} & \textbf{50.4} & \textbf{68.4} & \textbf{80.5} & \textbf{76.3} & \textbf{95.0} & \textbf{31.0} & \textbf{27.9} & \textbf{65.0} & \textbf{37.4} & \textbf{32.2} & \textbf{75.0} \\
\midrule
GRESO          & 83.2 & 80.8 & --    & \textbf{24.9} & 23.3 & --    & 51.6 & 48.5 & --    & \textbf{81.3} & 74.2 & --    & 29.0 & 25.4 & --    & 35.4 & 29.6 & --    \\
+Ours          & \textbf{83.5} & \textbf{81.2} & \textbf{80.0} & 25.1 & \textbf{23.4} & \textbf{100.0} & \textbf{53.3} & \textbf{49.9} & \textbf{73.7} & 80.3 & \textbf{74.6} & --    & \textbf{31.2} & \textbf{27.1} & \textbf{70.0} & \textbf{37.2} & \textbf{31.4} & \textbf{75.0} \\
\bottomrule
\end{tabular}%
}
\caption{Per-benchmark results on mathematical reasoning, Qwen3-4B. Best Acc ($\uparrow$), AUC ($\uparrow$), and S2B (\%, $\downarrow$) for each benchmark and method. Best results per column in \textbf{bold}.}
\label{tab:math_4b_full}
\end{table*}

%% file: section/Appendix/complete_res.tex
\section{Complete Experimental Results}
\label{appendix:complete_results}

\subsection{Rollout Dynamics Under Learnability-Aware Sampling}
\label{app:rollout_dynamics}

\begin{figure}[h]
    \centering
    \includegraphics[width=0.49\linewidth]{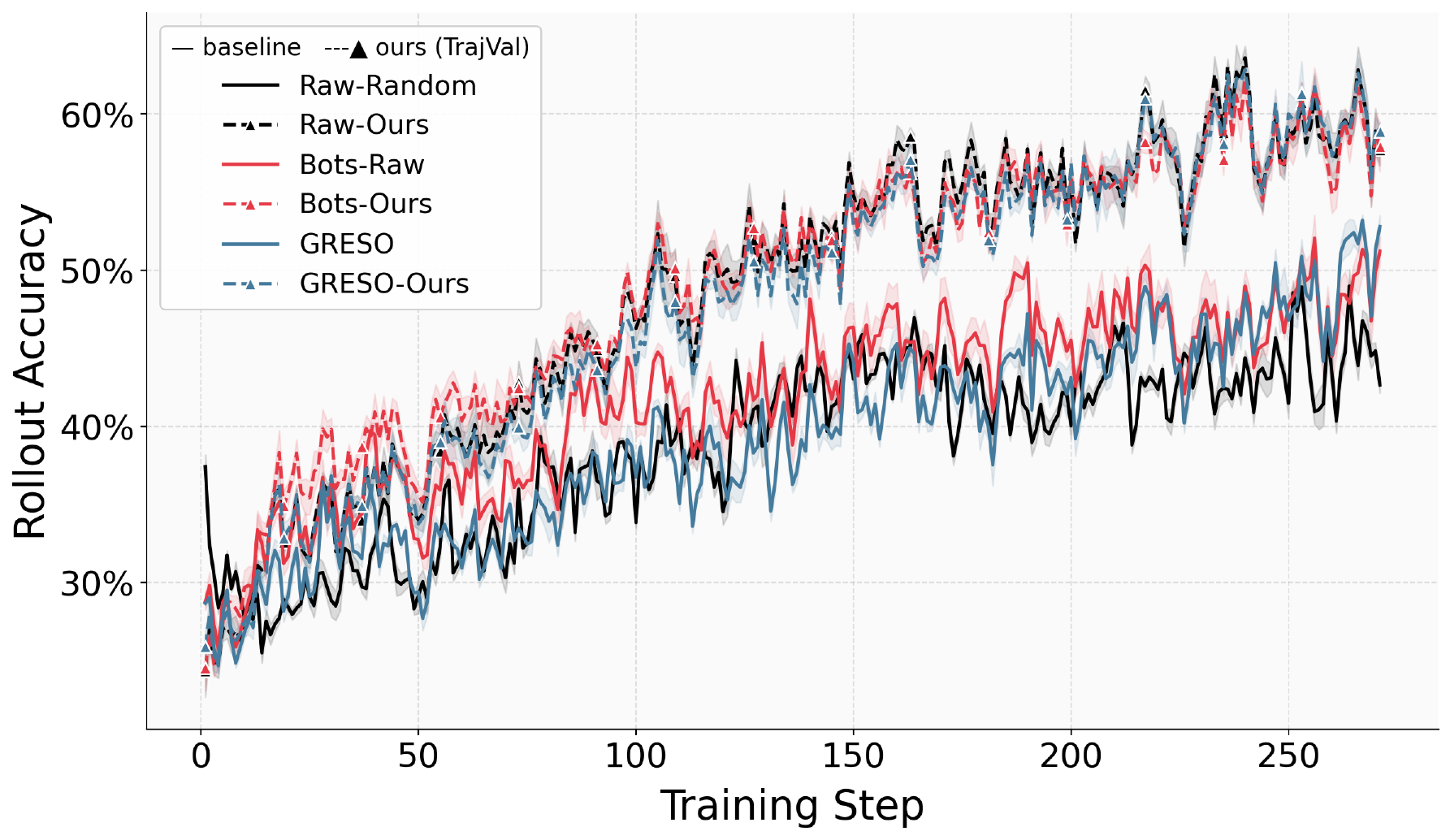}
    \hfill
    \includegraphics[width=0.49\linewidth]{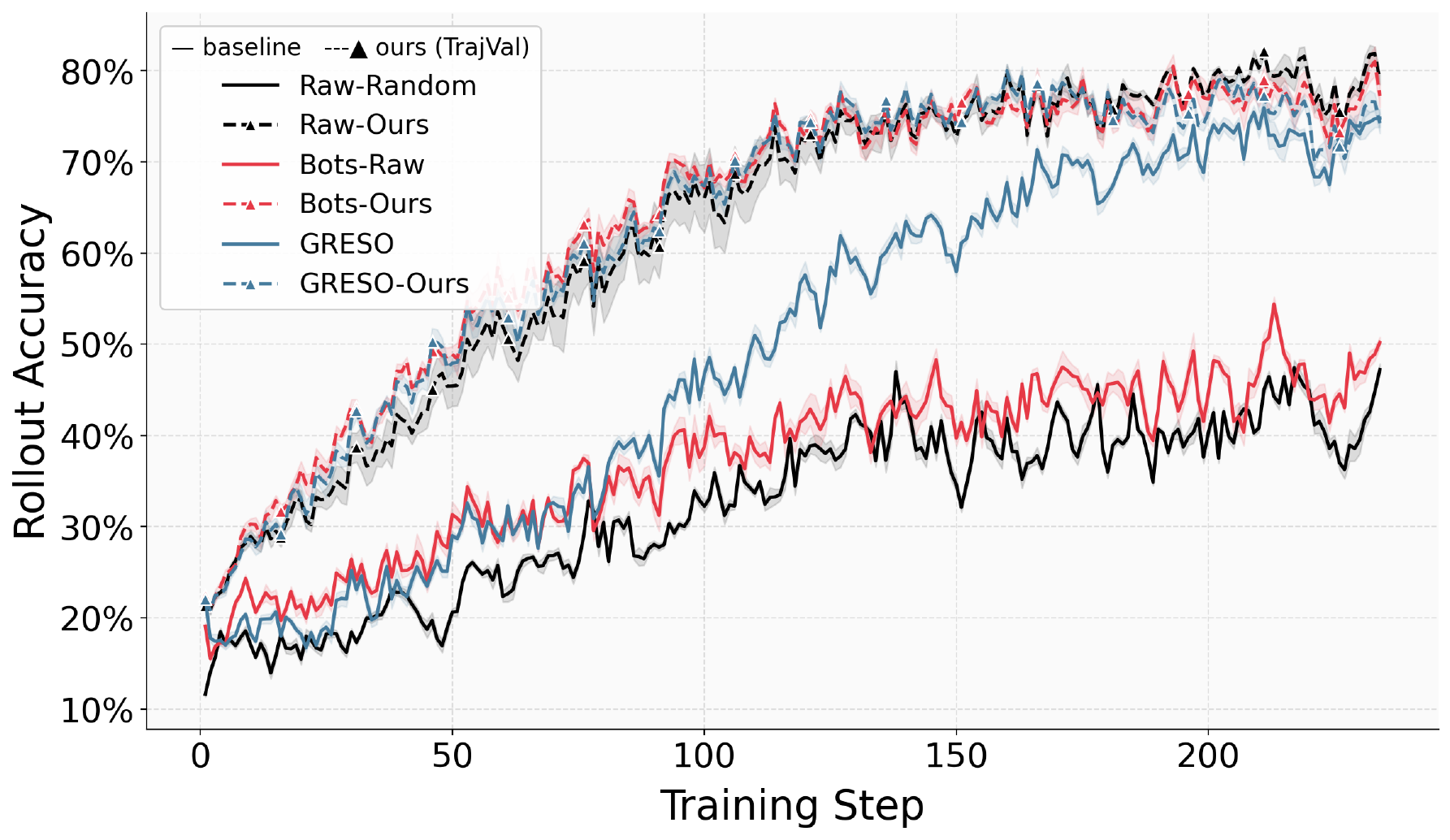}
    \caption{Training-time rollout accuracy under different sampling strategies for Qwen3-1.7B. Left: Math. Right: Logic. Solid lines denote the original baselines, and dashed lines denote their \textsc{TrajVal}-augmented counterparts. The plots provide a qualitative view of how learnability-aware sampling changes the training dynamics on the sampled pool.}
    \label{fig:appendix_rollout_dynamics}
\end{figure}

Figure~\ref{fig:appendix_rollout_dynamics} visualizes the training-time rollout accuracy under different sampling strategies for Qwen3-1.7B on Math and Logic. These curves are intended as a diagnostic view of how \textsc{TrajVal} reshapes the sampled training stream. They are \textbf{\emph{not}} a direct substitute for downstream benchmark performance, which is reported in Section~\ref{sec:main}.

\begin{itemize}[leftmargin=*]
    \item \textbf{\emph{Earlier rise in rollout accuracy.}} Across both domains, the \textsc{TrajVal}-augmented variants generally increase rollout accuracy faster than their corresponding baselines, with a clearer separation on Math. This pattern is consistent with the intended effect of learnability-aware sampling: allocating more updates to tasks whose rewards are more responsive to continued optimization.
    
    \item \textbf{\emph{Baseline-specific behavior is preserved early on.}} In the early stage of training, each augmented curve remains close to its underlying baseline, rather than collapsing to a common trajectory. This suggests that \textsc{TrajVal} acts as a mild prior on top of the original scheduler, instead of overriding its native sampling behavior.
    
    \item \textbf{\emph{Greater alignment later in training.}} As training proceeds, the augmented variants become more similar to one another than the original baselines do, especially on Math. We view this as qualitative evidence that different schedulers equipped with the same learnability prior are gradually steered toward a more similar set of high-potential samples.
\end{itemize}

The contrast between Math and Logic is also informative: the same qualitative trends appear in both domains, but the Logic curves are noisier and the separations are smaller. This may reflect greater heterogeneity in the logic training pool, making rollout-level improvements less smooth even when the sampling prior is beneficial.

\subsection{Evaluation Curves}

\begin{figure*}[t]
    \centering
    \includegraphics[width=0.245\linewidth]{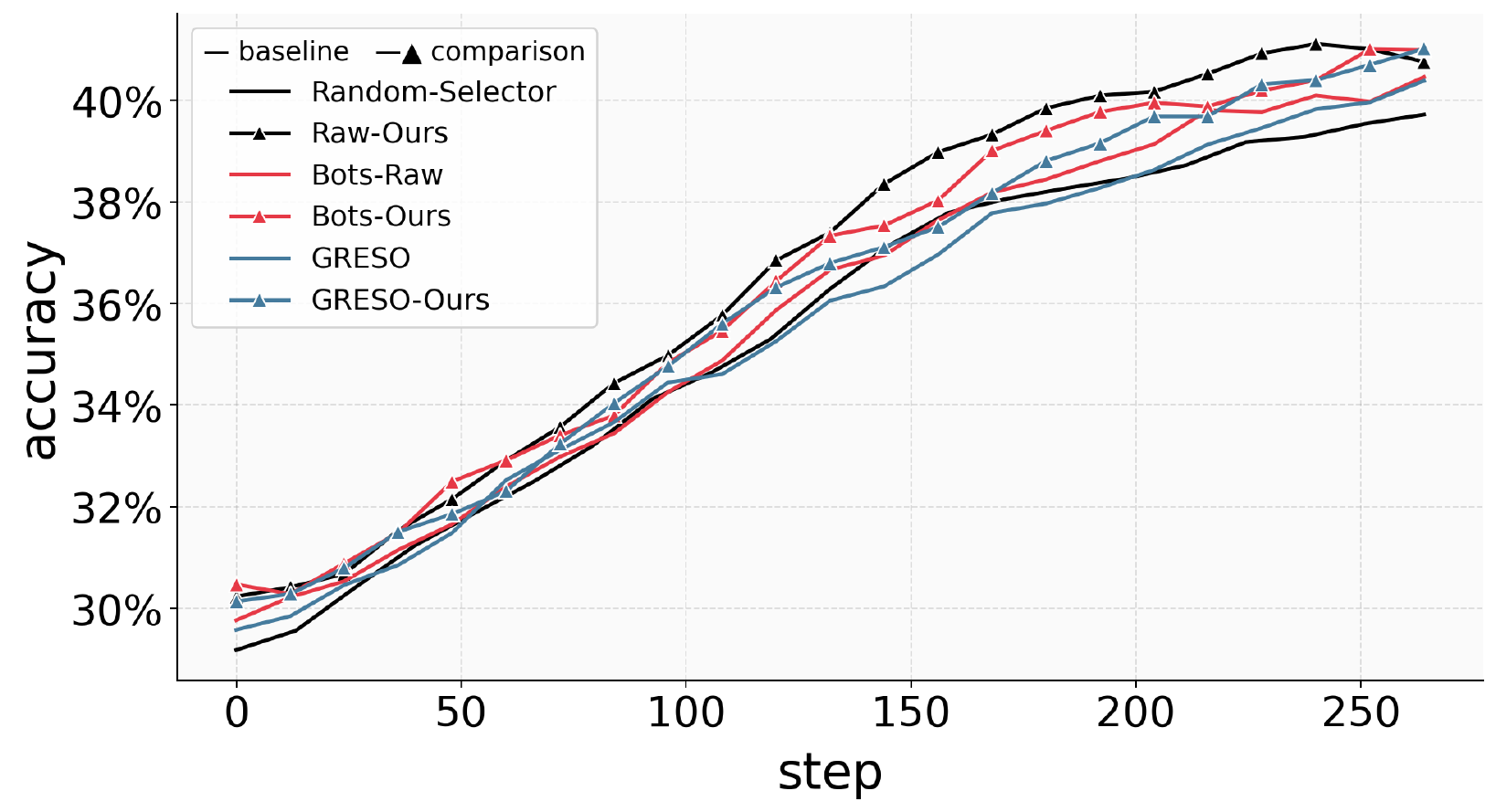}
    \hfill
    \includegraphics[width=0.245\linewidth]{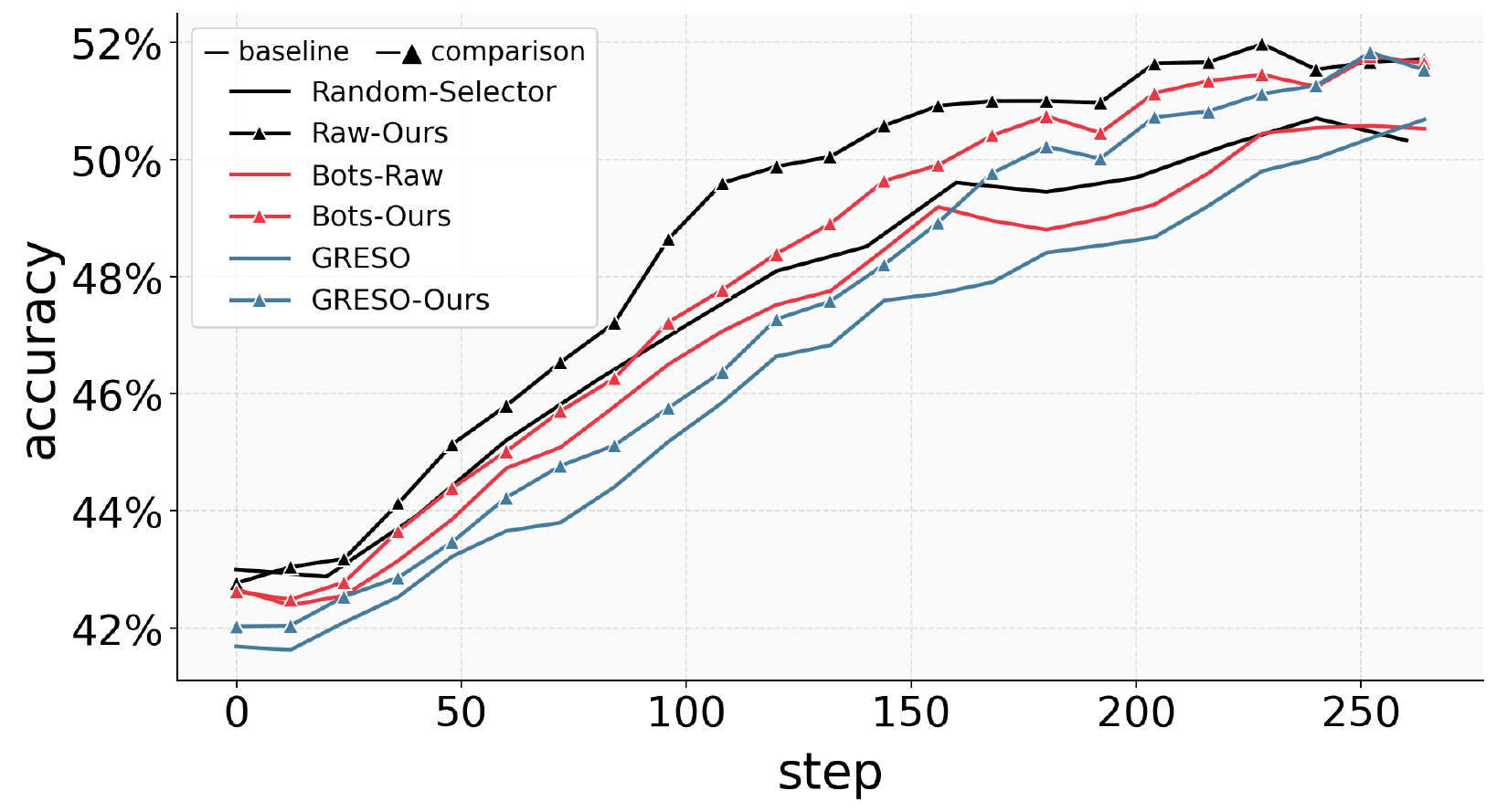}
    \hfill
    \includegraphics[width=0.245\linewidth]{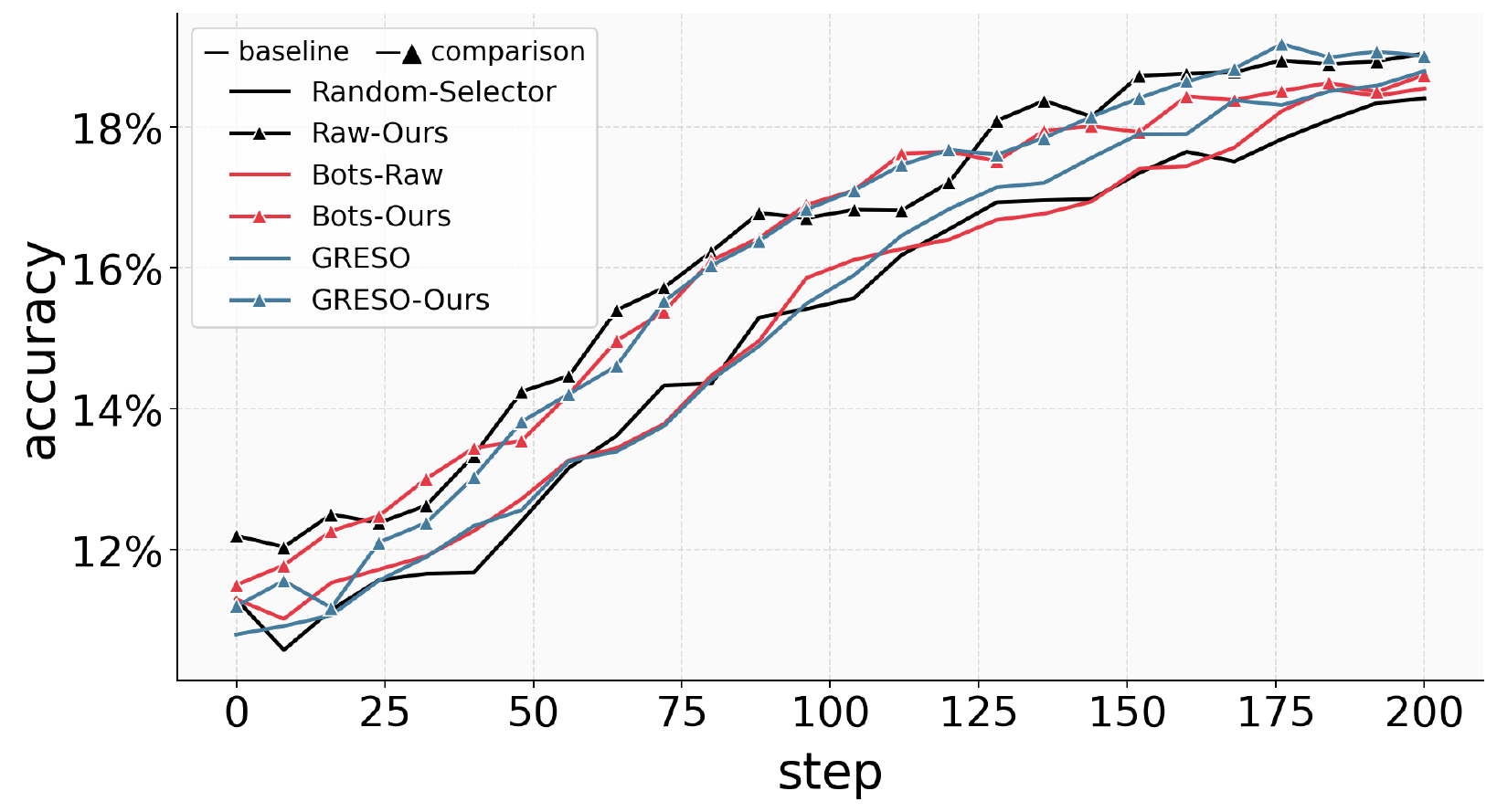}
    \hfill
    \includegraphics[width=0.245\linewidth]{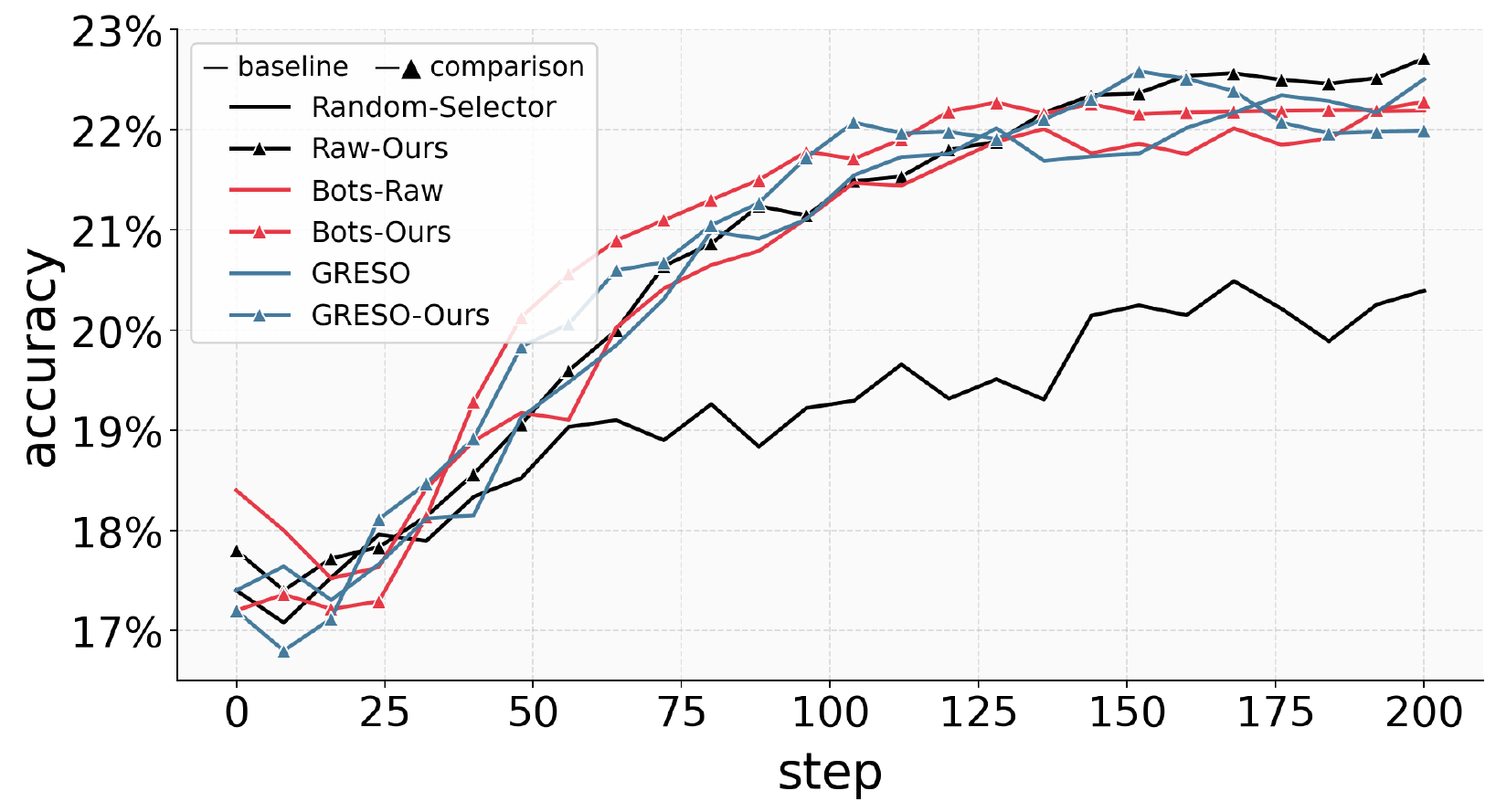}
    \caption{Evaluation accuracy curves for all main configurations on Math and Logic. The x-axis shows training steps and the y-axis shows evaluation accuracy.}
    \label{fig:eval_curves}
\end{figure*}

Figure~\ref{fig:eval_curves} presents the full evaluation trajectories for the main configurations across Math and Logic. Several consistent patterns are visible. First, \textsc{TrajVal}-augmented variants generally rise earlier than their corresponding baselines, indicating improved data efficiency in the early and middle stages of training. Second, these gains are not limited to faster initial progress: in most settings, the augmented curves also maintain a higher or comparable plateau, consistent with the improvements in Best Acc and AUC reported in Table~\ref{tab:main_result}. Third, the qualitative trend is shared across both standalone and plug-in use cases, suggesting that the learnability prior remains beneficial whether used on its own or combined with an online scheduler. Overall, the trajectory-level view reinforces the main conclusion that learnability-aware allocation improves training efficiency without sacrificing final evaluation performance.

\subsection{Complete Per-Benchmark Results}
\label{appendix:full_results}

Tables~\ref{tab:math_1b_full} and~\ref{tab:math_4b_full} report per-benchmark breakdowns of the main Math results for Qwen3-1.7B and Qwen3-4B, respectively. Each table reports Best Acc, AUC, and S2B for all six evaluation benchmarks (MATH500, Minerva, OlympiadBench, AMC23, AIME25, AIME24) under all three base training strategies and their \textsc{TrajVal}-augmented counterparts. Results on Llama-3.2-3B are reported separately in Appendix~\ref{app:stability_family}.

%% file: section/Appendix/overall_algorithm.tex
\section{Algorithm and Pseudocode of TrajVal}

Algorithm~\ref{algo:main} summarizes the full procedure.

\begin{algorithm*}[h]
\caption{\textsc{TrajVal}: Probe-Based Learnability Estimation for RL Post-Training}
\label{algo:main}
\begin{algorithmic}[1]

\State \textbf{Input:} training pool $\mathcal{D}$, base model $\theta_0$, probe size $|\mathcal{P}|$, probe epochs $T$, checkpoint window size $K$, compression exponent $\alpha$, floor $\epsilon$
\State \textbf{Output:} sampling distribution $\{p(x)\}_{x \in \mathcal{D}}$

\Statex

\vspace{0.5em}
\hrule
\vspace{0.5em}

\State \textit{// Phase I: estimate a learnability prior from a probe run}
\State Sample a probe set $\mathcal{P} \subset \mathcal{D}$ uniformly, with $|\mathcal{P}| \ll |\mathcal{D}|$
\State Train $\theta_0$ on $\mathcal{P}$ for $T$ epochs to obtain probe checkpoints $\{\theta_t\}$
\State Select early checkpoints $\mathcal{T}_{\mathrm{early}}$ and late checkpoints $\mathcal{T}_{\mathrm{late}}$, where $|\mathcal{T}_{\mathrm{early}}| = |\mathcal{T}_{\mathrm{late}}| = K$
\For{each task $x \in \mathcal{D}$}
    \State Compute endpoint estimates
    \[
    \bar{r}_{\mathrm{early}}(x) = \frac{1}{K}\sum_{t \in \mathcal{T}_{\mathrm{early}}} \mathrm{acc}(x,\theta_t),\qquad
    \bar{r}_{\mathrm{late}}(x) = \frac{1}{K}\sum_{t \in \mathcal{T}_{\mathrm{late}}} \mathrm{acc}(x,\theta_t)
    \]
\EndFor
\State $\{p(x)\}_{x \in \mathcal{D}} \gets \textsc{ComputeWeights}\bigl(\{\bar{r}_{\mathrm{early}}(x), \bar{r}_{\mathrm{late}}(x)\}_{x \in \mathcal{D}}, \alpha, \epsilon\bigr)$

\vspace{0.5em}
\hrule
\vspace{0.5em}

\Statex
\State \textit{// Phase II: main RL training with prior-weighted sampling}
\For{training step $t = 1, 2, \dots$}
    \State Sample a batch from $\mathcal{D}$ according to $p(x)$
    \Statex \hspace{\algorithmicindent}or use the composed weight $w_{\mathrm{final}}(x,t) = p(x)\, w_{\mathrm{online}}(x,t)$
    \State Perform one standard RL update on the sampled batch
\EndFor

\Statex
\Function{\textsc{ComputeWeights}}{$\{\bar{r}_{\mathrm{early}}(x), \bar{r}_{\mathrm{late}}(x)\}, \alpha, \epsilon$}
    \For{each task $x \in \mathcal{D}$}
        \State $\Delta(x) \gets \bar{r}_{\mathrm{late}}(x) - \bar{r}_{\mathrm{early}}(x)$
        \State $s(x) \gets \bigl(1 - \bar{r}_{\mathrm{early}}(x)\bigr)\max\!\bigl(0, \Delta(x)\bigr)$
        \State $w(x) \gets \max\!\bigl(s(x)^{\alpha}, \epsilon\bigr)$
    \EndFor
    \State \Return $p(x) = \dfrac{w(x)}{\sum_{x' \in \mathcal{D}} w(x')}$ for all $x \in \mathcal{D}$
\EndFunction

\end{algorithmic}
\end{algorithm*}